\documentclass[accepted]{uai2026} % after acceptance, for a revised version; 
\usepackage[american]{babel}
\usepackage{natbib} % has a nice set of citation styles and commands
    \renewcommand{\bibsection}{\subsubsection*{References}}
\usepackage{mathtools} % amsmath with fixes and additions
\usepackage{siunitx} % for proper typesetting of numbers and units
\usepackage{booktabs} % commands to create good-looking tables
\usepackage{tikz} % nice language for creating drawings and diagrams
 
\usepackage{amsmath,amsfonts,bm}

\def\eqref#1{equation~\ref{#1}}
\def\1{\bm{1}}

\DeclareMathAlphabet{\mathsfit}{\encodingdefault}{\sfdefault}{m}{sl}
\SetMathAlphabet{\mathsfit}{bold}{\encodingdefault}{\sfdefault}{bx}{n}

\def\gA{{\mathcal{A}}}

\def\gS{{\mathcal{S}}}

\def\gX{{\mathcal{X}}}

\usepackage{booktabs}
\usepackage{algorithm} 
\usepackage[noend]{algpseudocode}
\usepackage{tikz}
\usepackage{xcolor}
\usepackage{subcaption}
\usepackage{siunitx}
\usepackage{threeparttable}
\usepackage{adjustbox}    
\usepackage{rotating}
\usepackage{caption}
\usepackage{xcolor}
\usepackage{float}
\usepackage{booktabs}
\usepackage{threeparttable}
\usepackage{graphicx}   % for \resizebox
\usepackage{caption}
\usepackage{siunitx}
\usepackage{enumitem}
\usepackage{wrapfig} 
\usepackage{cleveref} 
\usepackage[table]{xcolor}
\usepackage{nicefrac}
\definecolor{lightgrayrow}{gray}{0.92}
 
\algrenewcommand{\algorithmiccomment}[1]{\hfill\textcolor{gray}{$\triangleright$ #1}}
\usepackage{pgfplots}

\title{Curriculum-Augmented GFlowNets For mRNA Sequence Generation}
 
\author[1]{Aya~Laajil}
\author[1]{Abduragim~Shtanchaev}
\author[1]{Aya~El~Mir}
\author[1]{Sajan~Muhammad}
\author[1]{Eric~Moulines}
\author[1]{Tiago~Silva}
\author[1]{Salem~Lahlou}
\affil[1]{%
    Mohamed bin Zayed University of Artificial Intelligence\\
    Abu Dhabi, United Arab Emirates
}

\begin{document}
 
\maketitle

\begin{abstract}

We develop an approach for the \emph{de novo} synthesis of messenger RNA (mRNA) sequences encoding target proteins.
This requires optimizing multiple biological objectives over large combinatorial spaces defined by synonymous codon sequences, a setting where Reinforcement Learning (RL) methods suffer from mode collapse and Generative Flow Networks (GFlowNets) face sparse rewards over long horizons. 
To address these challenges, we propose \textbf{Curriculum-Augmented GFlowNets (CAGFNs)}, a multi-objective GFlowNet trained with an adaptive length-based curriculum driven by learning progress.
In a nutshell, CAGFNs learn from progressively harder tasks corresponding to longer target proteins. 
We evaluate CAGFNs on \textbf{\texttt{CodonDesignEnv}}, a novel codon-level environment that preserves protein identity via dynamic action masking.
Our experiments show that CAGFNs achieve competitive or stronger Pareto-front coverage and diversity while converging up to \textbf{4 $\times$ faster} than non-curriculum GFlowNets, and maintaining \textbf{100\% sequence uniqueness} compared to RL baselines.\looseness=-1 

\end{abstract}

\begin{figure*}
    \centering
    \includegraphics[width=0.8\linewidth]{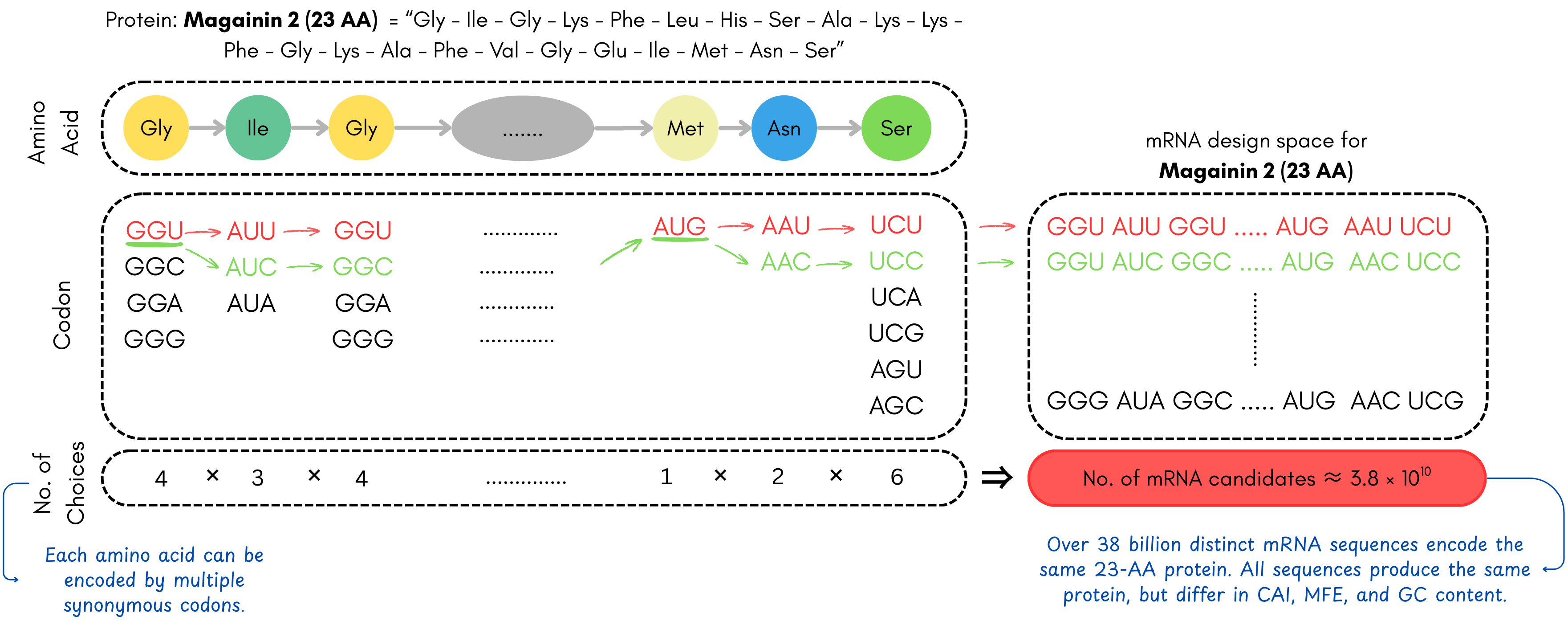}
    \caption{Codon redundancy in the mRNA design space. Each amino acid in Magainin 2 (23 amino acids) admits multiple synonymous codons; the product of choices across all positions yields 
    $\approx 3.8 \times 10^{10}$ distinct mRNA sequences encoding the same protein. Two example sequences are highlighted. All candidates produce the same protein but differ in biological properties such as stability, translation efficiency, and nucleotide composition. 
    }
    \label{fig:mrnadesignex}
\end{figure*}

\section{Introduction}
Imagine a molecule that can be designed to instruct human cells to produce a protein of interest. Such is the promise of messenger RNA (mRNA), which has become a cornerstone of modern biotechnology \citep{pardi2018mrna, sahin2014mrna}. Designing \textit{de novo} mRNA sequences that encode a target protein and achieve optimality on particular properties of interest \citep{gustafsson2004codon, kane1995effects, mauger2019mRNA} is therefore of growing practical importance. This task can be framed as generating long, structured sequences under multiple, often competing objectives, which makes search and optimization challenging \citep{keeney1993decisions, zhang2023algorithm, angermueller2020population}.

Because biological targets are diverse and downstream outcomes are difficult to predict, diversity is a central design criterion \citep{mullis2019diversity}. This need is further reinforced by the limited fidelity of early-stage screening methods, such as in silico simulations or in vitro assays, which only partially capture downstream biological outcomes. To maximize the likelihood that at least one candidate proves successful, it is therefore important that the proposed sequences broadly cover the different modes of the underlying ``goodness'' function that estimates future success. The search space in this scenario is large, and the best solutions often lie not at a single optimum but across a diverse Pareto front of trade-offs. 
Additionally, due to codon redundancy (see \Cref{ex:mrna_design}), the number of synonymous nucleotide sequences that encode a protein grows exponentially with its length, and those sequences differ dramatically in biological properties. For example, the SARS-CoV-2 spike protein ($\sim$1,273 amino acids) has on the order of $10^{632}$ possible synonymous sequences, most of which are suboptimal for practical use \citep{zhang2023algorithm}. Therefore, the goal of mRNA sequence design is to generate diverse sequences that (i) preserve the desired protein sequence, (ii) satisfy biological constraints, and (iii) reveal explicit trade-offs between competing objectives so practitioners can choose sequences suited to specific applications.
\looseness=-1 

Generative methods that can efficiently explore this combinatorial landscape, propose diverse high-quality candidates, and expose trade-offs between objectives are therefore of significant scientific and practical value. % \textcolor{gray}{Generative Flow Networks \citep[GFlowNets;][]{bengio2021flow} offer a principled framework for such tasks.}
Generative Flow Networks \citep[GFlowNets;][]{bengio2021flow} provide a diversity-aware sampling framework for such tasks. Unlike standard Reinforcement Learning (RL) or maximum likelihood methods, GFlowNets learn policies that sample complete objects $x$ with probability proportional to a user-specified non-negative reward $R(x)$. This property naturally supports the discovery of diverse high-reward solutions rather than collapse in a single mode, making them attractive for sequence design tasks where diversity is as important as optimization \citep{Jain2022_BioSeqGFlow,Roy2023_GoalCond}. Recent works have successfully applied GFlowNets to molecules and biological sequences \citep{Jain2022GFlowNet,zhu2023sample,Cretu2024SynFlowNet,koziarski2024rgfn}, showing their potential for structured generation in scientific domains.
\looseness=-1 

Despite these advantages and their recent success in scientific design tasks, training GFlowNets on a real-world sequence design problem such as mRNA reveals three fundamental challenges. First, sequences are long: rewards are only observed at the terminal step, leading to extremely sparse credit assignment. While GFlowNet objectives such as Trajectory Balance \citep{Malkin2022_TB} improve credit assignment over Flow Matching \citep{bengio2021flow} and Detailed Balance losses \citep{Bengio2023GFlowNet}, they are often difficult to optimize in long-horizon problems. Second, the design problem is inherently multi-objective: optimizing a fixed combination of objectives typically drives the search into a narrow region of the design space, obscuring alternative solutions that are valuable in practice. 
Third, and most importantly, each protein defines a distinct set of mRNA sequences, as dictated by the genetic code \citep{Crick1968}.
As a consequence, mRNA sequence generation exhibits a \emph{nested conditional structure}: each protein induces a distinct family of distributions over potential mRNA candidates, parameterized by our multi-objective reward function.
\looseness=-1 

Together, these challenges sparse credit assignment, multi-objective trade-offs, and protein specific structure motivate a training strategy that decomposes the problem by difficulty. Curriculum Learning (CL) \citep{bengio2009curriculum} offers a natural approach: the learner first masters simpler tasks before progressing to harder ones. Modern curricula go further by preferentially presenting tasks where the model is making the fastest progress, i.e., where the slope of the learning curve is largest. % For GFlowNets, 

We show that this principle significantly accelerates the training of GFlowNets in long-horizon, sparse-reward environments.

Concretely, we present an \emph{adaptive, length-based curriculum} that allocates training dynamically based on recent learning progress \citep{matiisen2019teacher}. 
To the best of our knowledge, this is the first study integrating adaptive curriculum learning with multi-objective GFlowNets with demonstrable empirical benefits. 
\looseness=-1

\paragraph{Contributions.}
We make the following contributions:
\begin{itemize}[leftmargin=12pt]
\item \textbf{mRNA Design Environment for GFlowNets.}
We formulate \textit{de novo} mRNA design as a multi-objective 
GFlowNet task over an exponentially large codon space, and 
introduce \texttt{CodonDesignEnv} which preserves protein identity 
via dynamic action masking.
\looseness=-1

\item \textbf{Curriculum-Augmented GFlowNets (CAGFN).}
We integrate adaptive, learning-progress–based curriculum learning with multi-objective GFlowNets, using protein length as a difficulty axis to stabilize and accelerate training in long-horizon settings.

\item \textbf{Empirical Validation.}
We show that CAGFN achieves competitive or stronger Pareto-front coverage and Top-$k$ diversity across protein lengths, reaching comparable performance \textbf{4$\times$ faster} than a long-only GFlowNet and maintaining \textbf{100\% uniqueness} compared to RL baselines (PPO \&  MOReinforce).
\looseness=-1 

\end{itemize}

\section{Related work}

Generative Flow Networks (GFlowNets) have been applied to a range of scientific
design problems, including molecules and biological sequences
\citep{bengio2021flow, Bengio2023GFlowNet, Jain2022GFlowNet}. Recent extensions address multi-objective optimization \citep{MOGFN2023} and synthesizability constraints \citep{Cretu2024SynFlowNet,koziarski2024rgfn}, highlighting their potential for structured generation tasks. 

To our knowledge, GFlowNets have not yet been applied to mRNA sequence design, which poses distinct challenges due to codon redundancy, biological constraints, and long sequences with sparse rewards. Curriculum learning has been shown to improve training in sparse-reward settings \citep{bengio2009curriculum, matiisen2019teacher}, but has not been combined with GFlowNets. 

In this work, we explore the integration of adaptive curriculum learning with multi-objective GFlowNets for mRNA sequence design. A more detailed related work is provided in \Cref{app:related-work}.

\begin{figure*}[h]
    \centering
    \includegraphics[width=0.8\linewidth]{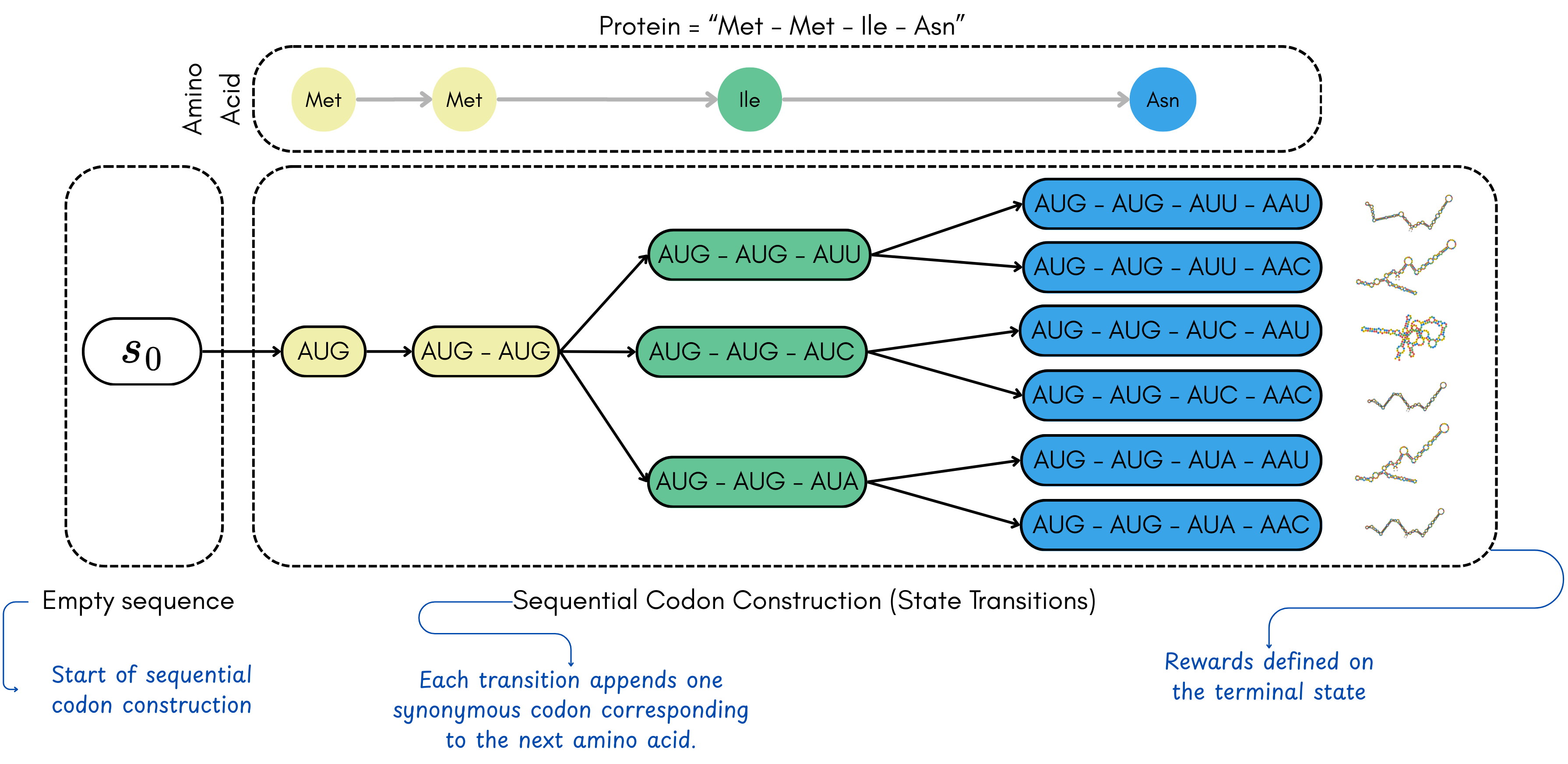}
    \caption{Tree representation of the mRNA sequence design process for the illustrative protein Met–Met–Ile–Asn. Each node corresponds to a codon selection step, and edges represent possible transitions.  For this example, Met has 1 codon (AUG), Ile has 3 (AUU, AUC, AUA), and Asn has 2 (AAU, AAC).  Terminal states correspond to mRNA sequences encoding this protein, each with unique biological properties.   This DAG (tree in this case) forms the basis for the GFlowNet to explore diverse and high-quality sequences. In particular, notice that each target protein induces a distinct DAG.
    % In particular, notice that each target protein induces a distinct DAG. % Terminal states correspond to mRNA sequences encoding this protein, each with unique biological properties. 
    % This DAG (tree in this case) forms the basis for the GFlowNet to explore diverse and high-quality sequences.
    }
    \label{fig:mRNA_DAG}
\end{figure*}

\section{Background}

\subsection{ \protect\lowercase{m}RNA sequence design}

Messenger RNA (mRNA) is a linear polymer built from a nucleotide alphabet $\Sigma=\{\text{A},\text{U},\text{G},\text{C}\}$. Proteins are sequences of amino acids and each amino acid is encoded by a \emph{codon}, a contiguous triplet of nucleotides. The standard genetic code maps the set of $4^{3} = 64$ codons onto 20 amino acids plus stop signals.
As several codons can encode the same amino acid, the mapping is \emph{degenerate} (see example in \Cref{fig:mrnadesignex}). 
We say two codons are \emph{synonyms} if they represent the same amino acid.  
This redundancy creates a large combinatorial set of possible mRNA sequences encoding the same protein, which necessitates algorithmic search over sequences that preserve protein identity while satisfying biological and structural constraints. \looseness=-1

\paragraph{Design objectives. }
\label{mrna_objectives}
To maximize the chances that at least one of the designed mRNAs will be successfully translated into the target protein, it is important to cover as much as possible the modes of a \textit{goodness} function estimating the candidate's translational efficiency. Diversity-aware methods, such as GFlowNets, combine breadth with targeted exploration of multiple promising modes, improving the probability of downstream success and accelerating the overall design cycle \citep{pyzer2018bayesian,terayama2021black}, see \Cref{diversity} for further discussion. \looseness=-1

Also, mRNA design is inherently multi-objective: beyond preserving the protein sequence, we aim to optimize biological properties such as translational efficiency, stability and protein expression \citep{sahin2014mrna}. Three commonly used objectives for mRNA design \citep{sahin2014mrna,pardi2018mrna} include:
\begin{enumerate}[leftmargin=12pt]
    \item \textbf{Codon Adaptation Index (CAI)}: A metric that quantifies how closely a coding sequence matches the preferred codon usage of a target species. A higher CAI indicates better alignment with species-specific codon preferences and is generally associated with improved translation efficiency \citep{Sharp1987Codon, Lee2018}.

    \item \textbf{Minimum Free Energy (MFE)}: The (negative) folding free energy of the mRNA's secondary structure. A lower MFE (more negative) indicates a more stable folded structure (see, e.g., \citet{Zuker1981Optimal}). \looseness=-1 
    \item \textbf{GC content}: The fraction of G or C nucleotides in the sequence. High GC-content generally correlates with increased mRNA stability and translational efficiency \citep{courel2019gc}, in particular, optimizing GC-content often increases the likelihood of successful protein synthesis \citep{zhang2023algorithm}. \looseness=-1 
\end{enumerate} 

% \vspace{-0.5cm}
\subsection{GFlowNets}
GFlowNets \citep{bengio2021flow, Bengio2023GFlowNet} are a class of probabilistic generative models that learn a stochastic policy to construct compositional objects \(x\in\mathcal{X}\) (e.g., graphs, sequences) by sampling trajectories \(\tau\) in a directed acyclic graph (DAG) \(G=(\mathcal{S},\mathcal{A})\). Construction begins at an initial empty state \(s_0\) and proceeds by iteratively sampling actions \(a\in\mathcal{A}\) according to a forward policy \(P_F(\cdot\mid s)\) until a terminal state \(x=s_n\) is reached. The aim is to learn \(P_F\) so that the marginal probability of sampling \(x\) under the forward policy is proportional to a given non-negative reward $R \colon \mathcal{X} \rightarrow \mathbb{R}_{+}$.
Denoting $\pi(x) = \frac{R(x)}{Z}$ as the target probability, where $Z \;=\; \sum_{x\in\mathcal{X}} R(x)$ is the partition function, the learning problem is to find parameters $\theta$ such that 
the induced marginal 
$\pi_\theta(x) = \sum_{\tau \rightsquigarrow x} P_{F,\theta}(\tau)$ 
approximates $\pi(x)$ for all $x \in \mathcal{X}$, when $\tau$ terminates at $x$.
\looseness=-1 

% \textcolor{orange}{where we write $\tau \rightsquigarrow x$ } 

\newcommand{\action}{u} % u is used in control; this avoids confusion with amino acids 

For a trajectory \(\tau=(s_0,\action_0,s_1,\dots,\action_{n-1},s_n=x)\) of states $\{s_{i}\}_{i=0}^{n}$ and actions $\{\action_{i}\}_{i=0}^{n-1}$, the GFlowNet forward and backward policies are defined by: $P_F(\tau)\;=\;\prod_{t=0}^{n-1} P_F(\action_t\mid s_t), \ P_B(\tau)\;=\;\prod_{t=1}^{n} P_B(\action_{t-1}\mid s_t).$ 

\looseness=-1 

In this work, we adopt the \emph{sub-trajectory balance} (Sub-TB) objective of \citet{Madan2023_SubTB} for policy learning. 
The reader may consult \citet{Malkin2022_TB, Madan2023_SubTB, bengio2021flow, Bengio2023GFlowNet} and \Cref{app:gflownets} for a more thorough introduction to GFlowNets. \looseness=-1 

\paragraph{Multi-objective (MO) GFlowNets.}
We recall basic terminology \citep{MOGFN2023}. 
Given objectives $R_{1}, \dots, R_{n}$ to be maximized, a point $x_{1} \in \mathcal{X}$ is dominated by $x_{2} \in \mathcal{X}$ if $R_{i}(x_{1}) \le R_{i}(x_{2})$ for all $i$ and $R_{j}(x_{1}) < R_{j}(x_{2})$ for at least one $j$. 
The Pareto front of a subset $\mathcal{K} \subseteq \mathcal{X}$ is defined as the set of non-dominated points in $\mathcal{K}$. 
 % MO GFlowNet, its 
 The \emph{Pareto performance} of a MO GFlowNet is the proportion of non-dominated sequences within a batch of samples. \looseness=-1

\section{GFlownets for \protect\lowercase{m}RNA Design}
 \label{sec:mrnageneration} 

We consider the problem of searching over a space of mRNA sequences encoding a target protein $a=(a_1,\dots,a_L)$ and finding diverse, high-quality mRNA sequences under a multi-objective reward function $R(x)$ capturing design objectives (\S~\ref{mrna_objectives}).
A codon sequence $x=(c_1,\dots,c_L)$ corresponds to an ordered sequence of codons where each codon $c_i \in \mathcal{C}(a_i)$ encodes an amino acid $a_i$. Here,  $\mathcal{C}(a_i)$ denotes the set of synonymous codons encoding the same amino acid $a_i$, which is derived from the genetic code. Thus, the feasible design space is $\mathcal{X} = \prod_{i=1}^L \mathcal{C}(a_i)$.
As most amino acids have multiple synonymous codons, $|\mathcal{X}|$ grows exponentially with $L$, as illustrated in \Cref{fig:mRNA_DAG} (In this setting, the DAG is a tree).
To investigate the capacity of GFlowNets for mRNA design, we introduce a new environment \texttt{CodonDesignEnv}. 
\looseness=-1

\paragraph{Environment, States, and Actions.}
The environment models mRNA sequence construction as a trajectory through states $s_t = (c_1,\ldots,c_t)$; $s_0$ is the empty sequence, and $s_t$ is a partially constructed mRNA sequence. 
A terminal state $x = s_L$ corresponds to a complete mRNA sequence encoding the target protein. 
The action space consists of the 64 codons plus a special exit action. 
Transitions are deterministic and constrained by the genetic code. 
At step $t<L$, the forward action set is dynamically masked to only allow synonymous codons encoding the $(t+1)$-th amino acid. 
At $t=L$, only the exit action is valid, forcing termination. 
% For backward sampling, only the last codon of the mRNA sequence can be removed, enforcing structured construction. 
These dynamic masks ensure that every trajectory from $s_0$ to the terminal sink state $s_f$ corresponds to a valid mRNA sequence, preserving the encoding of the original protein sequence. 
The resulting state-action graph is % a directed acyclic graph (DAG)~\ref{fig:mRNA_DAG} 
aligned with the codon redundancy of the genetic code, providing a natural fit for flow-based generative modeling with GFlowNets. 

Rewards are assigned only to terminal states and capture biologically motivated objectives (\S\ref{mrna_objectives}). For a complete mRNA sequence $x$, the reward is $R(x|w) = w^\top \phi(x)$,
where $\phi(x)$ is a vector of normalized objectives and $w \in \mathbb{R}^3$ are objective weights. 
By adjusting $w$, one can bias the GFlowNet toward different Pareto-optimal trade-offs, promoting diversity and mode coverage across the design space. 
Our framework supports both (i) unconditional training, where the weights of the reward are fixed and define a single trade-off between objectives, and (ii) conditional training, where the desired objective weights are provided at training and inference, enabling controlled exploration of the Pareto front.

\paragraph{Implicit protein conditioning.}
The environment is conditional on the target protein sequence $p$: both the reward function and valid action masks depend on $p$. However, the GFlowNet policy network does not receive an explicit protein embedding as input. 
Instead, conditioning arises implicitly through the dynamic action masking, which restricts available codons at each step based on the current amino acid. 
This design forces the model to learn generalizable codon selection principles rather than memorizing protein-specific solutions.
\looseness=-1

\paragraph{Unconditional Generation.}
In the unconditional setting, we fix the objective weight vector \(w\) during training so the GFlowNet learns a fixed reward function \(R(x \mid w)=w^\top\phi(x)\). 
The model is trained to assign flows over complete trajectories such that the sampling probability of the trajectory is proportional to its reward \(\pi_\theta(x)\propto R(x\mid w)\), encouraging the sampler to visit high-reward regions while preserving mode coverage. 
See Algorithm~\ref{alg:gfn_training} in \Cref{app:algorithms}.
At inference, we sample complete trajectories from the learned forward policy, producing a diverse pool of candidate mRNA sequences whose sampling probability reflects their composite reward under the chosen weights. This contrasts with optimization-only methods \citep{williams1992simple, schulman2017proximal, holland1992adaptation, snoek2012practical} that collapse to a single optimum.

    % The spread indicates exploration of diverse regions of the design space rather than mode collapse. 
    % Vertical lines denote the natural mRNA sequence scores.

\paragraph{Conditional Generation.}
As the biological objectives are often imperfect proxies for mRNA properties of interest, we aim to generate diverse Pareto-optimal solutions for each sub-problem \(\max_{x \in \mathcal{X}} R(x\mid w)\). 
Also, fixing a single set of weights for the GFlowNet % at a time 
prevents the model from exploiting the shared structure present between these related sub-problems.
\citet{bengio2021flow} extend standard GFlowNets to learn a single conditional policy that simultaneously models a family of distributions indexed by a conditioning variable. 
Each weight vector \( w \in \mathcal{W}\) induces a different reward function \(R(x\mid w)\) over terminal states \(x\in\mathcal{X}\). 
The conditioning information \(w\) is then available at every state.
\looseness=-1 

Denote by \(P_F(\cdot\mid s,w)\) the conditional forward policy, and let the conditional partition function be $Z(w)\;=\;\sum_{x\in\mathcal{X}} R(x\mid w)$.
The goal of a reward-conditional GFlowNet is to learn \(P_F(\cdot\mid s,w)\) (parameterized, e.g., by a neural network \(P_{F,\theta}(a\mid s,w)\)) such that the marginal probability of sampling a terminal state \(x\) under the forward policy satisfies \(\pi_\theta(x\mid w)\propto R(x\mid w)\). 
Exploiting the shared structure across \(w\)-conditioned reward functions allows a single conditional policy to model the entire family of sub-problems and, importantly, to generalize to different weight vectors \(w\). \looseness=-1

% --- paragraphs before the algorithm ---

% \begin{figure*}[h]
%     \centering
%     \includegraphics[width=0.8\linewidth]{metrics_with_bars.png}
%     \caption{\textbf{Distribution of reward metrics for unconditional mRNA generation with Multi-objective GFlowNets.} The dispersion of scores across generated sequences indicates exploration of multiple regions of the design space rather than concentration around a single mode. Vertical lines denote the natural mRNA sequence scores.}
%     \label{fig:metrics_distribution}
% \end{figure*}

%% MAIN ALGORITHM %% 
% \begin{minipage}{\linewidth}
\begin{algorithm}[t!]
\caption{Curriculum Learning Algorithm}
\label{alg:clgfn}
\begin{algorithmic}[1]
\small
\Require
% \Statex $T = \{t_1, \dots, t_N\}$  \Comment{Pool of tasks sets}
% \Statex $S = \{S_1, \dots, S_N\}$ \Comment{Pool of protein sequence sets for each $t_i$} 
% \Statex $P_{F,\theta}$ \Comment{Forward Policy (Student)}
% \Statex $\boldsymbol{w}_{\text{eval}}$  \Comment{Fixed weights for eval}
% \Statex $I_{\text{total}}$, $I_{\text{task}}$, $I_{\text{eval}}$, $\beta$, $\varepsilon$ \Comment{Hyperparameters}
% \State $H_k \leftarrow \emptyset$ for each task $t_k \in T$ \Comment{Initialize history for Learning Progress}
% \State $\mathrm{LP}_k \leftarrow 0$ for each task $t_k \in T$ \Comment{Initialize Learning Progress for $t_k$}
% \State $P \leftarrow \text{Uniform}(1/N)$

\Statex $T = \{t_1, \dots, t_N\}$ \Comment{Pool of task sets}
\Statex $S = \{S_1, \dots, S_N\}$ \Comment{Protein sequences for each $t_i$} 
\Statex $P_{F,\theta}$, $\boldsymbol{w}_{\text{eval}}$ \Comment{Forward policy \& eval weights}
\Statex $I_{\text{total}}$, $I_{\text{task}}$, $I_{\text{eval}}$, $\beta$, $\varepsilon$ \Comment{Hyperparameters}
\State $H_k \leftarrow \emptyset, \mathrm{LP}_k \leftarrow 0 \;\; \forall t_k \in T$ \Comment{Init history \& learning progress}
\State $P \leftarrow \text{Uniform}(1/N)$

\For{$i \in \{1, \dots, I_{\text{total}}\}$}
    \State $k \sim P$, $p_{\text{seq}} \sim S_k$   \Comment{Sample task \& protein}
    
    \State $\mathcal{E} \leftarrow \text{InitializeEnv}(p_{\text{seq}})$

    \For{$s  \in \{1, \dots, I_{\text{task}}\}$}     
        \State $\boldsymbol{w} \sim \text{Dirichlet}(\boldsymbol{\alpha})$
        \State $\mathcal{T} \sim P_{F,\theta}(\cdot | \mathcal{E}, \boldsymbol{w})$  \Comment{Sample trajectory}
        \State $\mathcal{L}_{\text{SubTB}} \leftarrow \text{CalculateLoss}(\mathcal{T})$
        \State $\theta \leftarrow \text{OptimizerStep}(\theta, \nabla_{\theta}\mathcal{L}_{\text{SubTB}})$ 
    \EndFor

    \If{$i \in \{I_{\text{eval}}, 2I_{\text{eval}}, \dots\}$}
       
        \For{each task $t_j \in T$}
            \State $p_{\text{seq}} \sim S_j$
            \State $\mathcal{E}_{\text{eval}} \leftarrow \text{InitializeEnv}(p_{\text{seq}})$
            \State $\mathcal{T}_{eval}$ $\sim P_{F,\theta}( \cdot | \mathcal{E}_{\text{eval}}, \boldsymbol{w}_{\text{eval}})$ \Comment{Eval trajectories} 
            % \State $r_j^{\text{new}} \leftarrow  
            % \sum_{\mathcal{T} \in\mathcal{T}_{eval}} R(\mathcal{T}) / |\mathcal{T}_{eval}| $ \Comment{Compute normalized reward}
            \State $r_j^{\text{new}} \leftarrow \frac{1}{|\mathcal{T}_{eval}|} \sum_{\mathcal{T} \in\mathcal{T}_{eval}} R(\mathcal{T})$ \Comment{Mean reward}
            % \State $r_j^{\text{prev}} \leftarrow \text{LastElementOf}(H_j)$ if $H_j$ is not empty else 0
            % \State $r_j^{\mathrm{prev}} \leftarrow \mathbb{1}\{\{|H_j|>0\}\}; H\{j,,|H_j|\}$
            \State $r_j^{\mathrm{prev}} \leftarrow \begin{cases} H_{j={|H_j|}}, & |H_j|>0 \\ 0 & |H_j|=0 \end{cases}$ 
            % \State $r_j^{\mathrm{prev}} \leftarrow \begin{cases} H_{j..|H_j|}, & |H_j|>0 $ \0$, & |H_j|=0 \end{cases}$
            \State $\Delta m_j \leftarrow r_j^{\text{new}} - r_j^{\text{prev}}$ 
            
            \State $\mathrm{LP}_j \leftarrow (1-\beta)\,\mathrm{LP}_j + \beta\,\Delta m_j$  \Comment{Update LP avg}
            \State $H_j \leftarrow H_j \cup \{r_j^{\text{new}}\}$  \Comment{Update the history}
        \EndFor
        
        \For{$j \in \{1, \dots, N\}$}
            \State $P(j) \leftarrow \max(0, \mathrm{LP}_j) + \varepsilon$ 
        \EndFor   
        \State $P \leftarrow P / \sum_{k=1}^N P(k)$ 
    \EndIf
\EndFor

\State \Return $P_{F,\theta}$

% \end{algorithmic}
% \end{wrapfigure}
\end{algorithmic}
\end{algorithm}
% \end{minipage}
%%%%%%%%%%%%%%%%%%%%%%%%%%%%

\section{Curriculum Learning}
\label{sec:curle}
Beyond weights generalization, we want the model to generalize across different types of proteins of different lengths. To do so, we introduce a \textbf{Curriculum-Augmented GFlowNets (CAGFN)}, an integration of Curriculum Learning (CL) with a multi-objective GFlowNet. 
The core idea is to expose a single, shared GFlowNet policy to training tasks of increasing difficulty, partitioned by protein length into intervals. The model incrementally acquires codon-level patterns on shorter proteins before progressing to longer, harder ones, allowing the same parameterized policy to refine and extend learned representations as task complexity grows. The curriculum advances adaptively based on learning progress, allocating training to intervals where the model improves most rapidly. We use the following task set ($\mathrm{T}$) of amino-acid (AA) length intervals in our experiments:
%The core idea is to expose the GFlowNet to multiple training tasks, so that the model incrementally acquires codon-level and local-structure patterns on different type of proteins. Practically, we partition training into length intervals and advance the curriculum according to a strategy that systematically adapts learning tasks of different difficulty based on the GFlowNet's learning progress. It allows the same parameterized policy to refine and extend learned representations as task complexity grows. This length-aware curriculum encourages hierarchical feature learning, improves sample efficiency, and reduces optimization instability when exploring extremely large combinatorial spaces. We use the following task set of amino-acid (AA) length intervals in our experiments:
% \begin{equation}
%     %\text{Tasks}  \;=\; \{[l_1,l_2],\;[l_3,l_4],\;[l_5,l_6],\;[l_7,l_8],\;[l_9,l_{10}]\}\ \text{(AA)},
%     \text{Tasks}  \;=\; \{[25,40],\;[45,60],\;[65,80],\;[85,120],\;[125,180]\}\ \text{(AA)}.
% \label{tasks}
% \end{equation}
\begin{equation}
\label{tasks}
%\text{Tasks}  \;=\; \{[l_1,l_2],\;[l_3,l_4],\;[l_5,l_6],\;[l_7,l_8],\;[l_9,l_{10}]\}\ \text{(AA)},
\mathrm{T} \;=\; 
\bigl\{[25,40],\;[45,60],\;[65,80], % \\
[85,120],\;[125,180]\bigr\}\;. % \text{(AA)}.
\end{equation}
Each pair %$(l_i, l_{i+1})$
defines an interval of protein lengths from which we sample during training. 
Given a sampled length, the corresponding protein sequence is then drawn from our pool of available proteins. At each step, we sample a task interval according to the LP-based distribution defined below (Eq.~\ref{eq:teacher_prob}).
\looseness=-1 

Curriculum Learning (CL) describes training procedures that expose a learner to a sequence of tasks of increasing difficulty \citep{bengio2009curriculum}. We use an automatic, adaptive curriculum where a ``Teacher'' selects tasks based on the ``Student's'' learning progress, rather than a hand-designed schedule \citep{matiisen2019teacher,willems2020mastering}.

\paragraph{Teacher–Student Curriculum Learning (TSCL).}
We follow the teacher-student (TSCL) paradigm \citep{matiisen2019teacher}, in which the Teacher maintains a per-task performance signal and uses \emph{learning progress} (LP) to bias sampling of tasks. 
For each task index $i\in\{1,\dots,K\}$ and discrete training step $t$ define a per-task metric $m_{i,t}$.
For example, we set $m_{i, t}$ to the average reward of the most valuable samples found at step $t$ for task $i$ in \Cref{sec:experiments}. 
The Teacher forms a smoothed learning-progress estimate $\mathrm{LP}_{i,t}$ from the discrete changes $\Delta m_{i,t}=m_{i,t}-m_{i,t-1}$. A common and stable choice for the learning-progress estimate $\mathrm{LP}_{i,t}$ is an exponential moving average: \looseness=-1 
\begin{equation}
\label{eq:lp_ema}
\mathrm{LP}_{i,t} \;=\; (1-\beta)\,\mathrm{LP}_{i,t-1} \;+\; \beta\,(m_{i,t}-m_{i,t-1}),
\end{equation}
with smoothing hyperparameter $\beta\in(0,1]$ (e.g. $\beta\approx 0.01\text{--}0.1$).  The Teacher then converts $\mathrm{LP}$ to a sampling distribution over tasks. A robust, positivity-enforcing rule is:
\looseness=-1 
\begin{equation}
\label{eq:teacher_prob}
P(i \mid t) \;=\; \frac{\max\!\big(0,\mathrm{LP}_{i,t}\big) + \varepsilon}{\sum_{j=1}^K\big(\max\!\big(0,\mathrm{LP}_{j,t}\big)+\varepsilon\big)},
\end{equation}
where $\varepsilon>0$ is a small floor to ensure exploration. We further describe and compare different curricula in \Cref{app:cl-configs}.

\paragraph{Choice of per-task metric.}
\label{sec:metrics}
The metric $m_{i,t}$ should reflect the Student's progress on task $i$. The design is robust to the specific metric as long as it (i) changes as the Student learns, and (ii) is reasonably smooth or smoothed by EMA to reduce noise. We choose for this work the  average reward over generated sequences, as a per-task metric, since it reflects the performance of our GFlowNets (the Student model).

\paragraph{$\mathrm{LP}$-based selection.} Selecting tasks with highest positive $\mathrm{LP}$ focuses training on regions where the Student still makes rapid gains, increasing sample efficiency. % and directing compute to the frontier of competence. 
Negative $\mathrm{LP}$ (decline in performance) can be used to trigger replay and reduce forgetting \citep{matiisen2019teacher}. 
% This approach necessitates a dynamic training setup where the environment's constraints change with each task. %We initialize a single, shared GFlowNet model. The central principle of our methodology is parameter sharing. The model does not learn a separate policy for each protein sequence; instead, it learns a single, conditional policy that can generalize across the entire task distribution.
% The model learns a single, conditional policy that generalizes across the entire task distribution, rather than memorizing protein-specific solutions. 
Importantly, we use a single multi-objective GFlowNet for every protein; only the dynamic action masking varies across tasks.  
This forces the model to learn the underlying principles of codon selection that lead to high rewards, rather than memorizing solutions for specific proteins. \looseness=-1 

The complete procedure is detailed in Algorithm~\ref{alg:clgfn}. In summary, it proceeds as follows:
\begin{itemize}[leftmargin=12pt]
    \item \textbf{Initialization:} We define the set of curriculum tasks $\mathrm{T}$ (cf.~Eq.\ref{tasks}), each associated with a pool of protein sequences. We created the pool of protein sequence sets for each task from the CodonTransformer dataset \citep{Fallahpour2025CodonTransformer} $\{S_i\}_{i=1}^{|\mathrm{T}|}$. 
    %The GFlowNet policy $\pi_\theta$, is initialized, along with the task-sampling distribution $P$, which initially assigns equal probability to all tasks.
    The GFlowNet policy $\pi_\theta$ is initialized, along with the task-sampling distribution $P$, which starts uniform but is \emph{dynamically updated} based on learning progress (Eq.~\ref{eq:teacher_prob}), progressively shifting probability toward tasks where the model improves fastest.

    \item \textbf{Task Sampling:} \label{sampling} At each training step, the Teacher samples a task index $k \sim P$. A protein sequence $p_{\text{seq}}$ is then drawn from the corresponding pool $S_k$, and a task-specific environment $\mathcal{E}$ is instantiated for that sequence  (Alg.~\ref{alg:clgfn}, lines 3–5). 
 
    \item \textbf{GFlowNet Training:} The shared policy $\pi_\theta$ is trained for $I_{\text{task}}$ iterations on this environment (Alg.~\ref{alg:clgfn},lines 6–11). At each iteration, a weight vector $\boldsymbol{w}$ is sampled, trajectories are generated, and the parameters $\theta$ are updated via the SubTB loss. Afterward, the environment is discarded, and a new task is sampled.

    \item \textbf{Evaluation and Adaptation:} At fixed intervals $I_{\text{eval}}$ (Alg.~\ref{alg:clgfn}, lines 12–28), the Student is evaluated on all tasks using a fixed evaluation weight vector $\boldsymbol{w}_{\text{eval}}$. For each task $t_j$, we compute the mean reward across generated sequences and compare it to the previous evaluation. The difference defines a learning progress signal $\Delta m_j$, which we use to update $\mathrm{LP}_j$. Tasks with higher $\mathrm{LP}_j$ are assigned greater sampling probability in the following training step, updating $P$ to favor tasks where the Student shows improvement.
\end{itemize}

\paragraph{Mapping to design criteria.}
Our framework addresses the three mRNA design criteria from Section~1: (i)~\emph{protein preservation} is guaranteed by dynamic action masking, which restricts codons to synonymous options; (ii)~\emph{biological constraints} are captured by CAI, MFE, and GC objectives; (iii)~\emph{trade-off exposure} is achieved via reward-conditional generation across the Pareto front.

\section{Experiments}
\label{sec:experiments}
All hyperparameter choices for curriculum, policy, and optimization are detailed in \Cref{appendix:hyperparams} (Tables~\ref{tab:curriculum-hparams}--\ref{tab:teacher-hparams}). Our code, based on the \texttt{torchgfn} library \citep{Lahlou2023TorchGFN} is available online\footnote{\url{https://anonymous.4open.science/r/GFN_for_mRNA_design-FCC3/}.}.
We begin our experiments by analyzing the model in an unconditional setting to establish a baseline, then proceed to the primary conditional generation task with and without CL.
\looseness=-1

\begin{figure}[t]
\centering
\begin{tikzpicture}
\begin{axis}[
    width=0.99\linewidth,
    height=0.55\linewidth,
    xlabel={Relative training time ($\downarrow{}$)},
    ylabel={Quality score $Q$ ($\uparrow$)},
    xmin=0, xmax=15,
    ymin=0.73, ymax=0.78,
    grid=both,
    legend style={at={(0.02,0.98)},anchor=north west,draw=none,fill=none},
    ticklabel style={font=\small},
    label style={font=\small},
]

% Points: (time, Q)
\addplot[only marks, mark=*, mark size=2.5pt] coordinates {(1, 0.7429)};
% \addlegendentry{SGFN (1x)}

\addplot[only marks, mark=square*, mark size=2.8pt] coordinates {(3, 0.7670)};
% \addlegendentry{CAGFN (3x)}

\addplot[only marks, mark=triangle*, mark size=2.8pt] coordinates {(7, 0.7572)};
% \addlegendentry{ROGFN (7x)}

\addplot[only marks, mark=diamond*, mark size=2.8pt] coordinates {(12, 0.7542)};
% \addlegendentry{LGFN (12x)}

% Optional labels next to points
\node[font=\small, anchor=west] at (axis cs:1,0.7429) {~SGFN};
\node[font=\small, anchor=west] at (axis cs:3,0.7670) {~CAGFN};
\node[font=\small, anchor=west] at (axis cs:7,0.7572) {~ROGFN};
\node[font=\small, anchor=west] at (axis cs:12,0.7542) {~LGFN};

\end{axis}
\end{tikzpicture}

\caption{\textbf{Quality-efficiency frontier.} Each point plots the relative training time versus a combined quality score 
$Q=\tfrac{1}{2}\left(\overline{\mathrm{TopKReward}}+\overline{\mathrm{TopKDiversity}}/41\right)$, ($K=50$). The diversity term is normalized by 41, which corresponds to the maximum achievable TopK diversity in our setting. Relative training time is reported with respect to SGFN. Averages are taken over all tasks in Table~\ref{tab:protein_all_metrics}.}

% \caption{\textbf{Quality--efficiency frontier.} Each point plots relative training time versus a combined quality score
% $Q=\tfrac{1}{2}\left(\overline{\mathrm{TopKReward}}+\overline{\mathrm{TopKDiversity}}/41\right)$, where averages are taken over all tasks in Table~\ref{tab:protein_all_metrics}.}

\label{fig:quality_efficiency_frontier}
\end{figure}

\paragraph{Preliminary Study.} 
Although our primary objective is conditional mRNA generation, we first conducted exploratory experiments in an unconditional setting to validate the core GFlowNet architecture. 
We used a MOGFN \citep{MOGFN2023} with fixed weights to generate mRNA sequences for a small protein sequence task. 
The model produced diverse sequences with competitive scores across key biological objectives, and to assess the validity of our generated sequences, we compared them with the corresponding naturally available mRNA sequence to check the range of our metrics. 
For example, the best generated sequence (vs. the natural one from our dataset in \Cref{dataset}) achieved a GC content of $53.78 \%$ (vs. $45.33 \%$), MFE of $-79.23$ (vs. $-68.20$), CAI of $0.63$ (vs. $0.54$), therefore outperforming the natural sequence in all metrics.  

% The Pareto front of the generated sequences is shown in Figure~\ref{fig:pareto}, while Figure~\ref{fig:metrics_distribution} illustrates the distributions of rewards of the generated sequences. 

However, as real-world mRNA design requires adapting to different weights configurations, focusing only on unconditional generation fails to capture the full Pareto front of biologically relevant solutions. In the remainder of this paper, we therefore focus exclusively on the conditional GFlowNet. \looseness=-1

We next assess the benefits of conditioning over the objectives' weights. 
To do so, given a target 35AA protein, we compared a multi-objective GFlowNet conditioned on the objectives weights and an unconditional GFlowNet with reward weights $(1/3, 1/3, 1/3)$ on a small protein sequence task.
For the conditional case, the reward weights are sampled during training from a $\mathrm{Dirichlet}(1,1,1)$ distribution.
Notably, Figure~\ref{fig:condi_vs_uncondi} shows that the conditional GFlowNet often finds more valuable states than its unconditional counterpart.
% illustrates the results, with a fixed objectives weigths (the reward is weighted sum of objectives given these weights). 

% \begin{figure}[t]
%     \centering

%     % (a) Pareto front — top
%     \begin{subfigure}[t]{\linewidth}
%         \centering
%         \includegraphics[width=0.45\linewidth]{pareto_front.png}
%         \caption{Pareto front.}
%         % \caption{Pareto front of generated sequences. Points on the frontier (circled) represent non-dominated trade-offs between MFE and GC content; interior points are dominated.}
%         \label{fig:pareto}
%     \end{subfigure}

%     \vspace{6pt}

%     % (b) Metrics distribution — bottom
%     \begin{subfigure}[t]{\linewidth}
%         \centering
%         \includegraphics[width=\linewidth]{metrics_with_bars.png}
%         \caption{Distribution of reward metrics.}
%         \label{fig:metrics_distribution}
%     \end{subfigure}

%     \caption{Unconditional mRNA sequence generation with GFlowNets under fixed objective weights $[0.3, 0.3, 0.4]$. 
%     \textbf{(a)} Pareto front. 
%     \textbf{(b)}~Distribution of reward metrics: The spread demonstrates that the model explores diverse regions of the design space rather than collapsing to a single mode. Vertical lines represent the natural sequence scores.}
%     \label{fig:unconditional_combined}
% \end{figure}

\paragraph{Comparison with RL.}
To evaluate its effectiveness, we compare our model against RL baselines for mRNA design. 
We consider the closely related MOReinforce \citep{lin2022pareto} as a baseline, as in % the MOGFN work 
\citep{MOGFN2023}. 
We did not compare our method with MARS \citep{xie2021mars} or standard MCMC approaches for mRNA design. MARS is designed for combinatorial optimization in small-molecule design, relying on iterative local mutations and chemical property predictors, which do not directly translate to the multi-objective nature of mRNA design. 
Similarly, traditional MCMC methods, while theoretically applicable, suffer from poor scalability in % extremely 
large combinatorial spaces % such as mRNA sequences 
and tend to produce low-diversity solutions concentrated in local modes \citep{terayama2021black, pyzer2018bayesian}. 
Instead, we compared our method to PPO \citep{schulman2017proximal}, an RL approach that has been successfully applied to sequence generation tasks. \looseness=-1

% CAGFN trades 0.10 reward points for complete sequence uniqueness and 63\% higher diversity.

\begin{table}[h]
    \centering
    \small
    \setlength{\tabcolsep}{3pt}
    \caption{
   CAGFN maintains 100\% uniqueness and greater diversity (measured as the average edit distance between the generated sequences) while achieving competitive reward. Metrics are averaged over 100 generated sequences. While MOReinforce achieves the highest scalar reward, it collapses to a \emph{single unique} sequence (1\% uniqueness), making it unsuitable for practical mRNA design where diversity is important.
}
    \looseness=-1 
    
    \begin{tabular}{lccc}
        \toprule
        Method & Reward ($\uparrow$) & Uniqueness (\%, $\uparrow$) & Diversity ($\uparrow$) \\
        \midrule
        MOReinforce & \textbf{0.69} (\scriptsize$\pm$0.05) & 1 & 19 \\
        PPO & 0.51 (\scriptsize$\pm$0.02) & 12 & 30 \\
        % MOGFN (w/o CL) & 0.59 (\scriptsize$\pm$0.064) & \textbf{100} & \textbf{31} \\
        \rowcolor{lightgrayrow} CAGFN & 0.59 (\scriptsize$\pm$0.066) & \textbf{100} & \textbf{31} \\
        \bottomrule
    \end{tabular}
    \label{tab:comparison}
\end{table}

As shown in Table~\ref{tab:comparison}, RL baselines like MOReinforce \citep{lin2022pareto} and PPO \citep{schulman2017proximal} suffered from mode collapse and low diversity, generating repetitive sequences (88\% duplicates for PPO). 
We measure diversity as the average edit distance between the generated sequences. Remarkably, although MOReinforce attains the highest scalar reward, it collapses to a single sequence.

Unlike MOReinforce, and consistent with \citet{MOGFN2023}, our GFlowNet approach achieves high rewards without collapsing to a single mode.

% \textcolor{red}{In contrast, as in \citep{MOGFN2023}, our GFlowNet approach maintained high rewards while generating fully unique and diverse candidates, highlighting its suitability for this exploration task.}
\looseness=-1

\subsection{Curriculum Learning Benefits}
To isolate the benefits of the CL strategy, we compare four distinct training strategies using the same conditional multi-objective GFlowNets (MOGFN) architecture: \looseness=-1
\label{variants}
\begin{itemize}[leftmargin=12pt]
    \item \textbf{Short-Only GFN (SGFN)}: A baseline model trained exclusively on a dataset of short protein sequences (30–60 amino acids). This tests the performance of a specialized model within its domain.
    \item \textbf{Long-Only GFN (LGFN)}: A second baseline trained exclusively on long protein sequences (125–180 amino acids). This model is used to assess the difficulty of learning complex, long-range dependencies from scratch.
    \item \textbf{Random-Order GFN (ROGFN)}: A control model trained using the full dataset of proteins spanning all length ranges. However, sequences are presented in a random, unstructured order. This baseline helps determine if performance gains come from data diversity alone, rather than the structured curriculum.
    \item \textbf{Curriculum-Augmented GFN (CAGFN) (Ours)}: The proposed model, trained on the full dataset using a curriculum that organizes tasks by increasing sequence length. This strategy is designed to enable the model to learn hierarchical patterns and generalize effectively.
\end{itemize}

\begin{table*}[t]
\centering
\scriptsize
\setlength{\tabcolsep}{3pt}
\begin{threeparttable}

\caption{\textbf{Comparison of generation quality across models and protein lengths.} We report TopK Reward and TopK Diversity (with $K=50$), as well as Pareto Performance, computed over 50 generated sequences per protein. Results are shown for five distinct protein targets (indexed 0--4) within each length range. Standard deviations across runs are reported in parentheses when available. Higher values indicate better performance for all metrics.}

% \caption{TopK Reward, TopK Diversity (K = 50), and Pareto Performance (over 50 generated sequences) across models and protein lengths. Standard deviations are shown in parentheses where available.}
\label{tab:protein_all_metrics}
\begin{tabular}{@{}l *{5}{c} *{5}{c} *{5}{c} @{}} 
\toprule
& \multicolumn{5}{c}{TopK Reward ($\uparrow$)} 
& \multicolumn{5}{c}{TopK Diversity ($\uparrow$)} 
& \multicolumn{5}{c}{Pareto Performance ($\uparrow$)} \\
\cmidrule(lr){2-6}\cmidrule(lr){7-11}\cmidrule(lr){12-16}
\textbf{Model}
 & 0 & 1 & 2 & 3 & 4
 & 0 & 1 & 2 & 3 & 4
 & 0 & 1 & 2 & 3 & 4 \\
\midrule

\multicolumn{16}{@{}l}{\textit{(25--60 AA)}} \smallskip\\
SGFN 
 & 0.58 \textcolor{gray}{(\scriptsize ±0.01)} & 0.52 \textcolor{gray}{(\scriptsize ±0.05)} & 0.55 \textcolor{gray}{(\scriptsize ±0.02)} & 0.64 \textcolor{gray}{(\scriptsize ±0.02)} & 0.67 \textcolor{gray}{(\scriptsize ±0.02)}
 & 24 \textcolor{gray}{(\scriptsize ±3.2)} & 32 \textcolor{gray}{(\scriptsize ±3.1)} & 37 \textcolor{gray}{(\scriptsize ±3.0)} & 40 \textcolor{gray}{(\scriptsize ±3.5)} & 36 \textcolor{gray}{(\scriptsize ±3.3)}
 & 0.08 & 0.06 & 0.09 & 0.16 & 0.19 \\

ROGFN
 & 0.58 \textcolor{gray}{(\scriptsize ±0.02)} & 0.53 \textcolor{gray}{(\scriptsize ±0.02)} & 0.55 \textcolor{gray}{(\scriptsize ±0.02)} & 0.67 \textcolor{gray}{(\scriptsize ±0.02)} & 0.70 \textcolor{gray}{(\scriptsize ±0.02)}
 & \textbf{25} \textcolor{gray}{(\scriptsize ±3.3)} & \textbf{32} \textcolor{gray}{(\scriptsize ±3.4)} & \textbf{38} \textcolor{gray}{(\scriptsize ±3.2)} & 40 \textcolor{gray}{(\scriptsize ±3.5)} & 37 \textcolor{gray}{(\scriptsize ±3.3)}
 & \textbf{0.13} & \textbf{0.15} & 0.17 & 0.17 & 0.21 \\

\rowcolor{lightgrayrow}
CAGFN
 & 0.61 \textcolor{gray}{(\scriptsize ±0.06)} & \textbf{0.58} \textcolor{gray}{(\scriptsize ±0.05)} & \textbf{0.62} \textcolor{gray}{(\scriptsize ±0.05)} & 0.69 \textcolor{gray}{(\scriptsize ±0.06)} & \textbf{0.73} \textcolor{gray}{(\scriptsize ±0.01)}
 & 25 \textcolor{gray}{(\scriptsize ±3.3)} & 32 \textcolor{gray}{(\scriptsize ±3.2)} & 38 \textcolor{gray}{(\scriptsize ±3.8)} & \textbf{41} \textcolor{gray}{(\scriptsize ±3.6)} & 37 \textcolor{gray}{(\scriptsize ±4.0)}
 & 0.07 & 0.12 & \textbf{0.21} & \textbf{0.21} & \textbf{0.22} \\

LGFN
 & \textbf{0.62} \textcolor{gray}{(\scriptsize ±0.03)} & 0.57 \textcolor{gray}{(\scriptsize ±0.03)} & 0.61 \textcolor{gray}{(\scriptsize ±0.02)} & \textbf{0.70} \textcolor{gray}{(\scriptsize ±0.03)} & \textbf{0.73} \textcolor{gray}{(\scriptsize ±0.03)}
 & 24 \textcolor{gray}{(\scriptsize ±3.1)} & 30 \textcolor{gray}{(\scriptsize ±3.2)} & 36 \textcolor{gray}{(\scriptsize ±3.3)} & 38 \textcolor{gray}{(\scriptsize ±3.5)} & 35 \textcolor{gray}{(\scriptsize ±3.3)}
 & 0.05 & 0.13 & 0.21 & 0.21 & 0.20 \\

\midrule
\multicolumn{16}{@{}l}{\textit{(85--120 AA)}} \smallskip\\
SGFN
 & 0.55 \textcolor{gray}{(\scriptsize ±0.03)} & 0.64 \textcolor{gray}{(\scriptsize ±0.02)} & 0.69 \textcolor{gray}{(\scriptsize ±0.02)} & 0.71 \textcolor{gray}{(\scriptsize ±0.02)} & 0.65 \textcolor{gray}{(\scriptsize ±0.03)}
 & 37 \textcolor{gray}{(\scriptsize ±3.0)} & \textbf{40} \textcolor{gray}{(\scriptsize ±3.2)} & \textbf{36} \textcolor{gray}{(\scriptsize ±3.7)} & 34 \textcolor{gray}{(\scriptsize ±3.4)} & 39 \textcolor{gray}{(\scriptsize ±3.1)}
 & 0.05 & 0.21 & 0.13 & 0.15 & 0.16 \\

ROGFN
 & 0.56 \textcolor{gray}{(\scriptsize ±0.02)} & 0.67 \textcolor{gray}{(\scriptsize ±0.02)} & 0.70 \textcolor{gray}{(\scriptsize ±0.02)} & \textbf{0.72} \textcolor{gray}{(\scriptsize ±0.02)} & 0.66 \textcolor{gray}{(\scriptsize ±0.02)}
 & 38 \textcolor{gray}{(\scriptsize ±3.4)} & \textbf{40} \textcolor{gray}{(\scriptsize ±3.3)} & \textbf{36} \textcolor{gray}{(\scriptsize ±3.2)} & \textbf{36} \textcolor{gray}{(\scriptsize ±3.7)} & 39 \textcolor{gray}{(\scriptsize ±3.0)}
 & 0.09 & 0.11 & \textbf{0.28} & 0.19 & \textbf{0.22} \\

\rowcolor{lightgrayrow}
CAGFN
 & 0.55 \textcolor{gray}{(\scriptsize ±0.06)} & 0.65 \textcolor{gray}{(\scriptsize ±0.06)} & 0.70 \textcolor{gray}{(\scriptsize ±0.05)} & 0.71 \textcolor{gray}{(\scriptsize ±0.08)} & \textbf{0.72} \textcolor{gray}{(\scriptsize ±0.01)}
 & 37 \textcolor{gray}{(\scriptsize ±3.2)} & \textbf{40} \textcolor{gray}{(\scriptsize ±3.3)} & \textbf{36} \textcolor{gray}{(\scriptsize ±3.9)} & 34 \textcolor{gray}{(\scriptsize ±3.5)} & \textbf{40} \textcolor{gray}{(\scriptsize ±3.2)}
 & \textbf{0.15} & \textbf{0.21} & 0.19 & \textbf{0.21} & 0.19 \\

LGFN
 & \textbf{0.60} \textcolor{gray}{(\scriptsize ±0.03)} & \textbf{0.70} \textcolor{gray}{(\scriptsize ±0.03)} & \textbf{0.73} \textcolor{gray}{(\scriptsize ±0.02)} & 0.69 \textcolor{gray}{(\scriptsize ±0.03)} & \textbf{0.72} \textcolor{gray}{(\scriptsize ±0.03)}
 & 37 \textcolor{gray}{(\scriptsize ±3.1)} & 38 \textcolor{gray}{(\scriptsize ±3.2)} & 34 \textcolor{gray}{(\scriptsize ±3.5)} & 35 \textcolor{gray}{(\scriptsize ±3.4)} & 38 \textcolor{gray}{(\scriptsize ±3.3)}
 & 0.05 & 0.13 & 0.13 & 0.21 & 0.20 \\

\bottomrule
\end{tabular}
\begin{tablenotes}
\item Indexes $0$-$4$ denote five distinct protein sequences sampled from each length range, ordered by increasing length within the range; see \Cref{tasks}.
\end{tablenotes}
\end{threeparttable}
\end{table*}

\paragraph{Loss Dynamics and Curriculum Progression. }
To better understand optimization behavior, we track both (i) the training loss and (ii) the distribution of protein sequence lengths sampled during training across the four regimes described in \S~\ref{variants}, and we report curriculum dynamics in \Cref{fig:loss_dynamics,fig:curriculum_overview} of \Cref{app:loss-curves}. 
Moreover, as shown in \Cref{fig:quality_efficiency_frontier}, CAGFN substantially reduces training time compared to both ROGFN and LGFN, while maintaining high-quality sequence generation. Concretely, CAGFN trains $4\times$ faster than LGFN and $2.4\times$ faster than ROGFN. This demonstrates that CL not only improves sample efficiency but also eases optimization when dealing with longer and more complex protein sequences. See Appendix~\ref{app:loss-curves} for the full set of loss curves.

To evaluate the effectiveness of our CL strategy, we test the trained models on unseen protein sequences from both the small (30-60 AA) and medium (85-120 AA) length ranges. We compare the CAGFN against the three baselines.

As shown in Table~\ref{tab:protein_all_metrics}, the results demonstrate that the CL approach yields a model that generates high-quality mRNA sequences and achieves a high Pareto performance. On the set of small proteins, the CAGFN achieves performance that is competitive with, and often superior to the specialist Short-Only GFN. This finding shows that training on a broad curriculum does not compromise performance on the simpler tasks learned early on. On medium-length proteins, CAGFN consistently matches or exceeds the performance of the other models. The comparison against the ROGFN is particularly revealing: while both models were trained on the same diverse dataset and generate high-quality sequences, the superior Pareto efficiency of CAGFN model indicates that the structured order of task presentation is a key ingredient. It allows the model to build hierarchical representations of codon selection patterns that generalize better than those learned from a random sequence of tasks.

This advantage is further illustrated in Figure~\ref{fig:quality_efficiency_frontier}, which presents the quality-efficiency frontier of all methods. Each model is plotted according to its relative training time and a combined quality score that aggregates TopK Reward and TopK Diversity across tasks. CAGFN lies on the Pareto frontier, achieving the highest overall quality while requiring substantially less training time than ROGFN and LGFN. Although SGFN trains slightly faster, it yields noticeably lower quality. These results confirm that the curriculum not only improves generation performance, but does so in a computationally efficient manner, offering the best trade-off between training cost and solution quality.

\section{Conclusion}
%\vspace{-0.5cm}
Motivated by the growing need for efficient and reliable mRNA therapeutics and vaccines, we addressed the problem of mRNA sequence design. The task focuses on optimizing and generating codon sequences that encode the same protein but differ in biological properties. We introduced Curriculum-Augmented Generative Flow Networks (CAGFN), a multi-objective GFlowNet enhanced with a teacher-driven curriculum. Building on a multi-objective GFlowNet, we designed a TSCL-style teacher that schedules tasks according to learning progress, enabling the model to generate diverse and high-quality mRNA sequences across proteins of varying lengths. 
Our experiments demonstrate that CAGFN outperforms non-curriculum GFlowNets, achieving higher rewards, broader Pareto-front coverage, and up to 4$\times$ faster convergence than LGFN. Against RL baselines, it maintains 100\% sequence uniqueness without sacrificing reward, generating diverse high-quality solutions without mode collapse.

\textbf{Limitations.}
Our optimization relies on proxies  that may not fully reflect \emph{in vivo} translational efficiency or protein expression, also wet-lab validation of the generated sequences remains an important next step. 
Additionally, our curriculum uses protein length as the sole difficulty axis, but other factors, such as amino acid composition or secondary structure complexity, may also affect learning difficulty. 

\textbf{Future work.} Natural extensions include multi-axis curricula (e.g., scheduling by both length and objective complexity), explicit protein conditioning via learned embeddings, and integration of more expressive reward functions such as learned oracles or experimental feedback loops. Scaling to full-length therapeutic proteins (>500 AA) and validating designs experimentally are key next steps toward practical deployment.

\newpage
% \section*{Reproducibility statement}
% To ensure complete reproducibility of our results, we provide implementation details and make all resources publicly available in \url{https://anonymous.4open.science/r/GFN_for_mRNA_design-FCC3/}. The complete source code includes the different training modes and the dataset we used for protein/mRNA task generation. The algorithms used are outlined in the appendices. The final set of hyperparameters used to reproduce the reported results are discussed in Appendix~\ref{appendix:hyperparams}, and provided as default values in our code. Additionally, we provide evaluation scripts for the comparison validation and comprehensive testing procedures that allow researchers to reproduce all experimental results reported in Section~\ref{sec:experiments}. 

\bibliography{uai2026-template/uai2026-template}

\begin{thebibliography}{39}
\providecommand{\natexlab}[1]{#1}
\providecommand{\url}[1]{\texttt{#1}}
\expandafter\ifx\csname urlstyle\endcsname\relax
  \providecommand{\doi}[1]{doi: #1}\else
  \providecommand{\doi}{doi: \begingroup \urlstyle{rm}\Url}\fi

\bibitem[Angermueller et~al.(2020)Angermueller, P{\"a}rnamaa, Parts, and Stegle]{angermueller2020population}
Christof Angermueller, Tanel P{\"a}rnamaa, Leopold Parts, and Oliver Stegle.
\newblock Population-based black-box optimization for biological sequence design.
\newblock In \emph{Proceedings of the 37th International Conference on Machine Learning (ICML)}, pages 52--62, 2020.

\bibitem[Bengio et~al.(2021)Bengio, Jain, Korablyov, Precup, and Bengio]{bengio2021flow}
Emmanuel Bengio, Moksh Jain, Maksym Korablyov, Doina Precup, and Yoshua Bengio.
\newblock Flow network based generative models for non-iterative diverse candidate generation.
\newblock \emph{Advances in Neural Information Processing Systems}, 34:\penalty0 27381--27394, 2021.

\bibitem[Bengio et~al.(2009)Bengio, Louradour, Collobert, and Weston]{bengio2009curriculum}
Yoshua Bengio, J{\'e}r{\^o}me Louradour, Ronan Collobert, and Jason Weston.
\newblock Curriculum learning.
\newblock In \emph{Proceedings of the 26th annual international conference on machine learning}, pages 41--48, 2009.

\bibitem[Bengio et~al.(2023)Bengio, Lahlou, Deleu, Hu, Tiwari, and Bengio]{Bengio2023GFlowNet}
Yoshua Bengio, Sami Lahlou, Thomas Deleu, Eugene~J. Hu, Mohit Tiwari, and Eric Bengio.
\newblock Gflownet foundations.
\newblock \emph{Journal of Machine Learning Research}, 24\penalty0 (210):\penalty0 1--55, 2023.

\bibitem[Courel et~al.(2019)Courel, Cl{\'e}ment, Bossevain, Foretek, Vidal~Cruchez, Yi, B{\'e}nard, Benassy, Kress, Vindry, Ernoult-Lange, Antoniewski, Morillon, Brest, Hubstenberger, Roest~Crollius, Standart, and Weil]{courel2019gc}
Marie Courel, Yohann Cl{\'e}ment, Camille Bossevain, Dominik Foretek, Olivier Vidal~Cruchez, Zheng Yi, Marion B{\'e}nard, Marie-No{\"e}lle Benassy, Micha{\"e}l Kress, Chlo{\'e} Vindry, Mich{\`e}le Ernoult-Lange, Christophe Antoniewski, Antonin Morillon, Patrick Brest, Alexis Hubstenberger, Hugues Roest~Crollius, Nancy Standart, and Dominique Weil.
\newblock Gc content shapes mrna storage and decay in human cells.
\newblock \emph{eLife}, 8, 2019.
\newblock \doi{10.7554/eLife.49708}.
\newblock URL \url{https://doi.org/10.7554/eLife.49708}.

\bibitem[Cretu et~al.(2024)Cretu, Harris, Roy, Bengio, and Li{\`o}]{Cretu2024SynFlowNet}
Miruna Cretu, Charles Harris, Julien Roy, Emmanuel Bengio, and Pietro Li{\`o}.
\newblock Synflownet: Towards molecule design with guaranteed synthesis pathways.
\newblock In \emph{ICLR 2024 Workshop on Generative and Experimental Perspectives for Biomolecular Design}, 2024.

\bibitem[Crick(1968)]{Crick1968}
F.~H.~C. Crick.
\newblock The origin of the genetic code.
\newblock \emph{Journal of Molecular Biology}, 38\penalty0 (3):\penalty0 367--379, 1968.
\newblock \doi{10.1016/0022-2836(68)90392-6}.

\bibitem[Fallahpour et~al.(2025)Fallahpour, Gureghian, Filion, Lindner, and Pandi]{Fallahpour2025CodonTransformer}
A.~Fallahpour, V.~Gureghian, G.~J. Filion, A.~B. Lindner, and A.~Pandi.
\newblock Codontransformer: a multispecies codon optimizer using context-aware neural networks.
\newblock \emph{Nature Communications}, 16\penalty0 (1):\penalty0 3205, 2025.
\newblock \doi{10.1038/s41467-025-XXXXX}.

\bibitem[Gustafsson et~al.(2004)Gustafsson, Govindarajan, and Minshull]{gustafsson2004codon}
Claes Gustafsson, Sridhar Govindarajan, and Jeremy Minshull.
\newblock Codon bias and heterologous protein expression.
\newblock \emph{Trends in biotechnology}, 22\penalty0 (7):\penalty0 346--353, 2004.

\bibitem[Holland(1992)]{holland1992adaptation}
John~H Holland.
\newblock \emph{Adaptation in natural and artificial systems: an introductory analysis with applications to biology, control, and artificial intelligence}.
\newblock MIT press, 1992.

\bibitem[Jain et~al.(2022{\natexlab{a}})Jain, Bengio, Hernandez-Garcia, Rector-Brooks, Dossou, Ekbote, ..., and Bengio]{Jain2022GFlowNet}
Meher Jain, Eric Bengio, Alejandro Hernandez-Garcia, Joshua Rector-Brooks, Benjamin~F. Dossou, Chaitanya~A. Ekbote, ..., and Yoshua Bengio.
\newblock Biological sequence design with gflownets.
\newblock In \emph{International Conference on Machine Learning}, pages 9786--9801. PMLR, June 2022{\natexlab{a}}.

\bibitem[Jain et~al.(2022{\natexlab{b}})Jain, Bengio, Hern\'{a}ndez-Garc{\'\i}a, Rector-Brooks, Dossou, Ekbote, Fu, Zhang, Kilgour, Zhang, Simine, Das, and Bengio]{Jain2022_BioSeqGFlow}
Moksh Jain, Emmanuel Bengio, Alex Hern\'{a}ndez-Garc{\'\i}a, Jarrid Rector-Brooks, Bonaventure F.~P. Dossou, Chanakya Ekbote, Jie Fu, Tianyu Zhang, Michael Kilgour, Dinghuai Zhang, Lena Simine, Payel Das, and Yoshua Bengio.
\newblock Biological sequence design with gflownets.
\newblock \emph{Proceedings of Machine Learning Research (PMLR), ICML 2022}, 162:\penalty0 9786--9801, 2022{\natexlab{b}}.
\newblock URL \url{https://proceedings.mlr.press/v162/jain22a.html}.

\bibitem[Jain et~al.(2023)Jain, Raparthy, Hern{\'a}ndez-Garc{\'\i}a, Rector-Brooks, Bengio, Miret, and Bengio]{MOGFN2023}
Moksh Jain, Sharath~Chandra Raparthy, Alex Hern{\'a}ndez-Garc{\'\i}a, Jarrid Rector-Brooks, Yoshua Bengio, Santiago Miret, and Emmanuel Bengio.
\newblock Multi-objective gflownets.
\newblock In \emph{International Conference on Machine Learning}, pages 14631--14653. PMLR, July 2023.

\bibitem[Kane(1995)]{kane1995effects}
James~F Kane.
\newblock Effects of rare codon clusters on high-level expression of heterologous proteins in escherichia coli.
\newblock \emph{Current opinion in biotechnology}, 6\penalty0 (5):\penalty0 494--500, 1995.

\bibitem[Keeney and Raiffa(1993)]{keeney1993decisions}
Ralph~L Keeney and Howard Raiffa.
\newblock \emph{Decisions with multiple objectives: preferences and value trade-offs}.
\newblock Cambridge university press, 1993.

\bibitem[Koziarski et~al.(2024)Koziarski, Rekesh, Shevchuk, van~der Sloot, Gai{\'n}ski, Bengio, Liu, Tyers, and Batey]{koziarski2024rgfn}
Micha{\l} Koziarski, Andrei Rekesh, Dmytro Shevchuk, Almer van~der Sloot, Piotr Gai{\'n}ski, Yoshua Bengio, Chenghao Liu, Mike Tyers, and Robert Batey.
\newblock Rgfn: Synthesizable molecular generation using gflownets.
\newblock \emph{Advances in Neural Information Processing Systems}, 37:\penalty0 46908--46955, 2024.

\bibitem[Kumar et~al.(2010)Kumar, Packer, and Koller]{kumar2010self}
M~Kumar, Benjamin Packer, and Daphne Koller.
\newblock Self-paced learning for latent variable models.
\newblock \emph{Advances in neural information processing systems}, 23, 2010.

\bibitem[Lee(2018)]{Lee2018}
Benjamin~D. Lee.
\newblock Python implementation of codon adaptation index.
\newblock \emph{Journal of Open Source Software}, 3\penalty0 (30):\penalty0 905, oct 2018.
\newblock \doi{10.21105/joss.00905}.
\newblock URL \url{https://doi.org/10.21105/joss.00905}.

\bibitem[Lin et~al.(2022)Lin, Yang, and Zhang]{lin2022pareto}
Xi~Lin, Zhiyuan Yang, and Qingfu Zhang.
\newblock Pareto set learning for neural multi-objective combinatorial optimization.
\newblock \emph{arXiv preprint arXiv:2203.15386}, 2022.

\bibitem[Madan et~al.(2023)Madan, Rector-Brooks, Korablyov, Bengio, Jain, Nica, Bosc, Bengio, and Malkin]{Madan2023_SubTB}
Kanika Madan, Jarrid Rector-Brooks, Maksym Korablyov, Emmanuel Bengio, Moksh Jain, Andrei~Cristian Nica, Tom Bosc, Yoshua Bengio, and Nikolay Malkin.
\newblock Learning gflownets from partial episodes for improved convergence and stability.
\newblock In \emph{International Conference on Machine Learning}, pages 23467--23483. PMLR, 2023.

\bibitem[Malkin et~al.(2022)Malkin, Jain, Bengio, Sun, and Bengio]{Malkin2022_TB}
Nikolay Malkin, Moksh Jain, Emmanuel Bengio, Chen Sun, and Yoshua Bengio.
\newblock Trajectory balance: Improved credit assignment in gflownets.
\newblock In \emph{Advances in Neural Information Processing Systems (NeurIPS 2022)}, 2022.
\newblock URL \url{https://arxiv.org/abs/2201.13259}.

\bibitem[Matiisen et~al.(2019)Matiisen, Oliver, Cohen, and Schulman]{matiisen2019teacher}
Tambet Matiisen, Avital Oliver, Taco Cohen, and John Schulman.
\newblock Teacher--student curriculum learning.
\newblock \emph{IEEE Transactions on Neural Networks and Learning Systems}, 31\penalty0 (9):\penalty0 3732--3740, 2019.

\bibitem[Mauger et~al.(2019)Mauger, Cabral, Presnyak, Su, Reid, Goodman, Link, Khatwani, Reynders, Esposito, et~al.]{mauger2019mRNA}
David~M Mauger, Benjamin~J Cabral, Valeria Presnyak, Shengdong Su, Daniel~W Reid, Ben Goodman, Kevin Link, Nikhil Khatwani, Jean-Paul Reynders, Daniel Esposito, et~al.
\newblock mrna structure regulates protein expression through stability and translation efficiency.
\newblock \emph{Nature Structural \& Molecular Biology}, 26\penalty0 (10):\penalty0 1089--1097, 2019.

\bibitem[Mullis et~al.(2019)Mullis, Rambo, Baker, and Reese]{mullis2019diversity}
Megan~M Mullis, Ian~M Rambo, Brett~J Baker, and Brandi~Kiel Reese.
\newblock Diversity, ecology, and prevalence of antimicrobials in nature.
\newblock \emph{Frontiers in microbiology}, 10:\penalty0 2518, 2019.

\bibitem[Pardi et~al.(2018)Pardi, Hogan, Porter, and Weissman]{pardi2018mrna}
Norbert Pardi, Matthew~J Hogan, Frederick~W Porter, and Drew Weissman.
\newblock mrna vaccines — a new era in vaccinology.
\newblock \emph{Nature Reviews Drug Discovery}, 17\penalty0 (4):\penalty0 261--279, 2018.

\bibitem[Pyzer-Knapp(2018)]{pyzer2018bayesian}
Edward~O Pyzer-Knapp.
\newblock Bayesian optimization for accelerated drug discovery.
\newblock \emph{IBM Journal of Research and Development}, 62\penalty0 (6):\penalty0 2--1, 2018.

\bibitem[Roy et~al.(2023)Roy, Bacon, Pal, and Bengio]{Roy2023_GoalCond}
Julien Roy, Pierre-Luc Bacon, Christopher Pal, and Emmanuel Bengio.
\newblock Goal-conditioned gflownets for controllable multi-objective molecular design.
\newblock \emph{arXiv preprint}, 2023.
\newblock URL \url{https://arxiv.org/abs/2306.04620}.

\bibitem[Sahin et~al.(2014)Sahin, Karik{\'o}, and T{\"u}reci]{sahin2014mrna}
Ugur Sahin, Katalin Karik{\'o}, and Ozlem T{\"u}reci.
\newblock mrna-based therapeutics — developing a new class of drugs.
\newblock \emph{Nature Reviews Drug Discovery}, 13\penalty0 (10):\penalty0 759--780, 2014.

\bibitem[Schulman et~al.(2017)Schulman, Wolski, Dhariwal, Radford, and Klimov]{schulman2017proximal}
John Schulman, Filip Wolski, Prafulla Dhariwal, Alec Radford, and Oleg Klimov.
\newblock Proximal policy optimization algorithms.
\newblock \emph{arXiv preprint arXiv:1707.06347}, 2017.

\bibitem[Sharp and Li(1987)]{Sharp1987Codon}
P.~M. Sharp and W.~H. Li.
\newblock The codon adaptation index-a measure of directional synonymous codon usage bias, and its potential applications.
\newblock \emph{Nucleic acids research}, 15\penalty0 (3):\penalty0 1281--1295, 1987.

\bibitem[Snoek et~al.(2012)Snoek, Larochelle, and Adams]{snoek2012practical}
Jasper Snoek, Hugo Larochelle, and Ryan~P Adams.
\newblock Practical bayesian optimization of machine learning algorithms.
\newblock \emph{Advances in neural information processing systems}, 25, 2012.

\bibitem[Terayama et~al.(2021)Terayama, Sumita, Tamura, and Tsuda]{terayama2021black}
Kei Terayama, Masato Sumita, Ryo Tamura, and Koji Tsuda.
\newblock Black-box optimization for automated discovery.
\newblock \emph{Accounts of Chemical Research}, 54\penalty0 (6):\penalty0 1334--1346, 2021.

\bibitem[Viviano et~al.(2023)Viviano, Younis, Choi, Schmidt, Bengio, and Lahlou]{Lahlou2023TorchGFN}
Joseph~D Viviano, Omar~G Younis, Sanghyeok Choi, Victor Schmidt, Yoshua Bengio, and Salem Lahlou.
\newblock torchgfn: A pytorch gflownet library.
\newblock \emph{arXiv e-prints}, pages arXiv--2305, 2023.

\bibitem[Willems et~al.(2020)Willems, Lahlou, and Bengio]{willems2020mastering}
Lucas Willems, Salem Lahlou, and Yoshua Bengio.
\newblock Mastering rate based curriculum learning.
\newblock \emph{arXiv preprint arXiv: 2008.06456}, 2020.

\bibitem[Williams(1992)]{williams1992simple}
Ronald~J Williams.
\newblock Simple statistical gradient-following algorithms for connectionist reinforcement learning.
\newblock \emph{Machine learning}, 8\penalty0 (3):\penalty0 229--256, 1992.

\bibitem[Xie et~al.(2021)Xie, Shi, Zhou, Yang, Zhang, Yu, and Li]{xie2021mars}
Yutong Xie, Chence Shi, Hao Zhou, Yuwei Yang, Weinan Zhang, Yong Yu, and Lei Li.
\newblock Mars: Markov molecular sampling for multi-objective drug discovery.
\newblock \emph{arXiv preprint arXiv:2103.10432}, 2021.

\bibitem[Zhang et~al.(2023)Zhang, Zhang, Lin, Xu, Li, Liu, Liu, Ma, Zhao, Jiang, et~al.]{zhang2023algorithm}
He~Zhang, Liang Zhang, Ang Lin, Congcong Xu, Ziyu Li, Kaibo Liu, Boxiang Liu, Xiaopin Ma, Fanfan Zhao, Huiling Jiang, et~al.
\newblock Algorithm for optimized mrna design improves stability and immunogenicity.
\newblock \emph{Nature}, 621\penalty0 (7978):\penalty0 396--403, 2023.

\bibitem[Zhu et~al.(2023)Zhu, Wu, Hu, Yan, Hou, Wu, et~al.]{zhu2023sample}
Yiheng Zhu, Jialu Wu, Chaowen Hu, Jiahuan Yan, Tingjun Hou, Jian Wu, et~al.
\newblock Sample-efficient multi-objective molecular optimization with gflownets.
\newblock \emph{Advances in Neural Information Processing Systems}, 36:\penalty0 79667--79684, 2023.

\bibitem[Zuker and Stiegler(1981)]{Zuker1981Optimal}
M.~Zuker and P.~Stiegler.
\newblock Optimal computer folding of large rna sequences using thermodynamics and auxiliary information.
\newblock \emph{Nucleic acids research}, 9\penalty0 (1):\penalty0 133--148, 1981.

\end{thebibliography}
% \bibliographystyle{icml2026}

%%%%%%%%%%%%%%%%%%%%%%%%%%%%%%%%%%%%%%%%%%%%%%%%%%%%%%%%%
%%%%%%%%%%%%%%%%% APENDINX %%%%%%%%%%%%%%%%% 
%%%%%%%%%%%%%%%%%%%%%%%%%%%%%%%%%%%%%%%%%%%%%%%%%%%%%%%%%
\clearpage
\onecolumn
\appendix
\crefalias{section}{appendix} % Tells cleveref to treat 'section' as 'appendix' from here on
\section{mRNA Design Problem}
\label{ex:mrna_design}
\paragraph{Codon Redundancy. }
The genetic code is redundant: multiple codons can encode the same amino acid. 
For example, the amino acid \textbf{Leucine (Leu)} is represented by six codons (UUA, UUG, CUU, CUC, CUA, CUG), while \textbf{glycine} has four (GGU, GGC, GGA, GGG). This allows different mRNA sequences to yield the same protein. As an illustration, a short peptide Met--Leu--Gly may be encoded by the mRNA sequence \texttt{AUG--UUA--GGU} or alternatively by \texttt{AUG--CUG--GGC}, both producing the identical protein despite differing at the nucleotide level. Exploiting codon redundancy is central to mRNA sequence design, as it provides a vast combinatorial space in which sequences can be optimized for multiple objectives while preserving the encoded protein.

To clarify the nature of the mRNA design problem, we provide a simple example that highlights the role of codon redundancy and the search space of feasible mRNA sequences. 

\paragraph{Example : A short protein sequence. }  
Consider designing an mRNA sequence for a short protein consisting of three amino acids: Methionine–Phenylalanine–Lysine \textbf{(Met–Phe–Lys)}.  
\begin{itemize}
    \item Met (M) can only be encoded by the codon \texttt{AUG}.  
    \item Phe (F) can be encoded by \texttt{UUU} or \texttt{UUC}.  
    \item Lys (K) can be encoded by \texttt{AAA} or \texttt{AAG}.  
\end{itemize}
The feasible design space for this protein is therefore:

\[
\begin{aligned}
\{ &\texttt{AUG UUU AAA},\; \texttt{AUG UUU AAG}, \\
   &\texttt{AUG UUC AAA},\; \texttt{AUG UUC AAG} \}.
\end{aligned}
\]

All four codons sequences produce the same protein, but differ in codon usage patterns. 
Choosing among them involves evaluating biological properties such as GC content, minimum free energy (MFE), and codon adaptation index (CAI). 
This simple case illustrates the exponential growth of the design space as protein length increases, underscoring the need for generative models like GFlowNets to explore it effectively. 

\section{Motivation for diversity}
\label{diversity}
Diversity is essential in biological sequence design because targets, mechanisms of action, and structural contexts are highly heterogeneous and often change over time \citep{mullis2019diversity}. Cheap preclinical screens (in silico or in vitro) are imperfect proxies for eventual performance in wet-lab experiments, so concentrating effort on a single 'best' sequence risks failure if that sequence aligns with biases or blind spots of the surrogate assays. By contrast, proposing a diverse set of high-quality candidates increases the chance that at least one design occupies a different mode of the true, and unknown, success landscape, providing robustness to model misspecification and experimental noise. Practically, this is critical when searching astronomical, combinatorial spaces (e.g., on the order of $10^{60}$ possible sequences) under tight experimental budgets: diversity-aware methods combine broad coverage with targeted exploration of multiple promising modes, improving the probability of downstream success and accelerating the overall design cycle \citep{pyzer2018bayesian,terayama2021black}. Generative models that can exploit the combinatorial structure in these spaces have the potential to speed up the design process for such sequences.

\section{Background on GFlowNets}
\label{app:gflownets}
GFlowNets (GFNs) are a family of probabilistic models that learn a stochastic policy to generate compositional objects, such as sequences or molecular graphs, through a sequence of construction steps, with probability proportional to their reward $R(x)\ge 0$ \citep{bengio2021flow}. 

We consider a directed acyclic graph $(\gS,\gA)$, where nodes $\gS$ represent states and edges $\gA$ represent actions. If $(s,s')\in \gA$, we say that $s$ is a parent of $s'$ and that $s'$ is a child of $s$. There is a unique state $s_0$ with no parents that we name \textit{source state} and a unique state $s_f$ with no children that we name \textit{sink state}. We refer to the parents of the \textit{sink state} as terminating states $\gX = \{ s \in \gS \mid (s,s_f) \in \gA)\}$. A complete trajectory $\tau = (s_0, s_1,...,s_n,s_{n+1} = s_f)$ is a sequence of states that start with the \textit{source state} and ends with the \textit{sink state}.

We also consider a reward function $R : \gX \rightarrow \mathbf{R}$ such that $\forall x \in \gX, R(x) > 0$. By convention, we can also consider that the reward for non terminating states is zero $\forall s \in \gS \setminus \gX, R(s) = 0$. The goal of the GFlowNet is to learn to sample terminating states proportionally to the reward function $\pi(x) \propto R(x)$. To achieve this, GFlowNets start from the initial state (for example, empty sequence or molecule in biological setting, initial coordinates in hyper-grid environment...) and iteratively sample actions to move to the following states. 

Formally, we consider a non-negative flow function $F: \gA \rightarrow \mathbf{R}$. The goal of the GFlowNet framework is to learn such a flow function that satisfies the following flow matching (FM) and reward matching constraints, for all $s' \neq s_0,s_f$ and for all $x \in \gX$, respectively:
\begin{align}
    &\sum_{(s,s')\in\gA} F(s \rightarrow s') = \sum_{(s',s'')\in\gA} F(s' \rightarrow s''),\\
    &F(x \rightarrow s_f) = R(x).
\end{align}
We extend the definition of the flow function to states:
\begin{align}
    \forall s \in \gS,\quad F(s) = \sum \limits_{(s,s')\in\gA} F(s \rightarrow s').
\end{align}
We also define the forward and backward policies as follow:
\begin{align}
    P_F(s'\mid s) = \frac{F(s \rightarrow s')}{F(s)}, \ \
    P_B(s\mid s') = \frac{F(s \rightarrow s')}{F(s')}.
\end{align}

\paragraph{Trajectory Balance (TB). } Given a complete trajectory/episode $\tau=(s_0\!\to\! s_1\!\to\!\dots\!\to\! s_n=x)$ that terminates in $x$, the trajectory balance objective enforces a global balance using a learned normalizing constant $Z\!>\!0$ \citep{Malkin2022_TB}:
\begin{equation}
\label{eq:tb}
\mathcal{L}_{\text{TB}}(\tau)
\;=\; \left(\log\frac{Z\prod_{t=0}^{n-1} P_F(s_{t+1}\!\mid\! s_t)}{R(x)\prod_{t=0}^{n-1} P_B(s_t\!\mid\! s_{t+1})}\right)^{\!2}.
\end{equation}
Minimizing this loss across sampled trajectories ensures that $Z\prod_t P_F(\cdot)\approx R(x)\prod_t P_B(\cdot)$, which leads to sampling proportional to $R(x)$ under appropriate parameterizations.

\paragraph{Sub-trajectory TB (SubTB). } To enable learning from partial episodes, SubTB \citep{Madan2023_SubTB} applies a TB-like constraint to partial trajectories $\tau_{0:k}=(s_0,\dots,s_k)$:
\begin{equation}
\label{eq:subtb}
\mathcal{L}_{\text{SubTB}}(\tau_{i:k})
\;=\; \left(\log\frac{F(s_i)\prod_{t=i}^{k-1} P_F(s_{t+1}\!\mid\! s_t)}{F(s_k)\prod_{t=i}^{k-1} P_B(s_t\!\mid\! s_{t+1})}\right)^{\!2},
\end{equation}
where $F(s)$ is the learned flow at state $s$. 

% In this work, we conducted our experiments using TB and SubTB losses for training GFlowNets. SubTB is particularly well-suited for our curriculum learning approach, as it naturally accommodates training on sequences of varying lengths and improves convergence for long mRNA sequences compared to standard TB loss.

During development, we compared TB and SubTB losses and found that TB fails to converge for sequences longer than approximately 55 amino acids (see Figure~\ref{fig:loss_comparison} in Appendix~\ref{app:loss-curves}). All reported experiments therefore use the SubTB objective, which naturally accommodates sequences of varying lengths and is well-suited to our curriculum learning approach.

% \section{GFlowNet Training Loop Algorithm}
% \label{app:gfn_algorithm}
\subsection{Training Algorithm}

\label{app:algorithms} We provide the detailed pseudocode for the GFlowNet training loop. Algorithm~\ref{alg:gfn_training} is the basis of Algorithm~\ref{alg:clgfn}, which outlines the teacher-student interaction, task sampling, and learning progress updates.

\begin{minipage}{\linewidth}
\begin{algorithm}[H]
\caption{GFlowNet Training Loop}
\label{alg:gfn_training}
\begin{algorithmic}[1]
\Require 
$\mathcal{E}$ (env), $P_{F,\theta}$ (forward policy), $R(\cdot)$ (reward), $B$ (batch), $I$ (iters),
$\mathcal{O}$ (opt), $\boldsymbol{\alpha}$

\For{$i \in \{1, \dots, I_{\text{total}}\}$}
    \State $\boldsymbol{w} \gets \textsc{None}$
    \If{conditional}
        \State $\boldsymbol{w} \sim \mathrm{Dirichlet}(\boldsymbol{\alpha})$
    \EndIf
    \State $\mathcal{T}_{1:B} \sim P_{F,\theta}(\cdot \mid \mathcal{E}, \boldsymbol{w})$ \Comment{Sample $B$ paths}
    \State $r_{1:B} \gets R(\mathcal{T}_{1:B})$
    \State $\mathcal{L} \gets \mathrm{Loss}(\mathcal{E}, \mathcal{T}_{1:B}, r_{1:B})$ \Comment{Eq.~\ref{eq:subtb}}
    \State $\theta \gets \theta - \mathcal{O}\!\big(\nabla_{\theta}\mathcal{L}\big)$
\EndFor
\State \Return $P_{F,\theta}$
\end{algorithmic}
\end{algorithm}
\end{minipage}

\section{Hyperparameters and implementation details}
\label{appendix:hyperparams}
\subsection{Reward Details}
In our implementation, we made use of previously published repositories, making small adaptations where necessary. We used the torchgfn library \citep{Lahlou2023TorchGFN} to implement the GFlowNet model.
To compute reward, we used 3 objectives : the \textbf{Codon Adaptation Index (CAI)}, a measure of how well the sequence conforms to species-specific codon usage preferences. Higher CAI typically leads to more efficient translation \citep{Sharp1987Codon}. To compute the CAI of mRNA sequences, we used the CAI library \citep{Lee2018}. The second objective is the \textbf{Minimum Free Energy (MFE)}, a (negative) folding free energy of the mRNA's secondary structure. A lower MFE (more negative) indicates a more stable folded structure. Then, third one is the \textbf{GC content,} which is the fraction of G or C nucleotides in the sequence. High GC-content generally correlates with increased mRNA stability and translational efficiency \citep{courel2019gc}, in particular, optimizing GC-content has a similar effect to optimizing codon usage \citep{zhang2023algorithm}.

\subsection{Dataset}
\label{dataset}
We base our experiments on the dataset provided by the CodonTransformer paper \citep{Fallahpour2025CodonTransformer}, which contains natural mRNA sequences corresponding to a diverse set of proteins. Each entry includes a DNA sequence, the corresponding protein sequence, and gene and organism information. Since our task focuses on mRNA design, we converted the DNA sequences into their corresponding mRNA sequences before using them in our experiments.

\subsection{Hyperparameters}
This appendix lists the hyperparameters and implementation choices used for curriculum learning and the policy function (PF) in our GFlowNet experiments. Values shown are the defaults used in all experiments reported in the paper. We also briefly summarise the alternative ranges explored during tuning and the selection rationale.

\begin{table}[htbp]
  \centering
  \caption{Curriculum learning hyperparameters}
  \label{tab:curriculum-hparams}
  \small
  \begin{tabular}{@{}ll@{}}
    \toprule
    Hyperparameter & Value \\
    \midrule
   \texttt{curriculum\_tasks} &
[25,40], [45,60], [65,80], \\
& [85,120], [125,180] \\

    \texttt{n\_iterations} & 100 \\
    \texttt{eval\_every} & 5 \\
    \texttt{train\_steps\_per\_task} & 200 \\
    \bottomrule
  \end{tabular}
\end{table}

\vspace{3pt}

\vspace{3pt}
\paragraph{Policy function (PF). }
We choose for the PF module in the GFlowNets a Transformer architecture with the following hyperparameters: \(\texttt{embedding\_dim}=32\), \(\texttt{hidden\_dim}=256\), \(\texttt{n\_hidden}=4\) Transformer layers, \(\texttt{n\_head}=8\).

\begin{table}[htbp]
  \centering
  \caption{Optimization and sampling hyperparameters}
  \label{tab:optim-hparams}
  \small
  \begin{tabular}{@{}ll@{}}
    \toprule
    Hyperparameter & Value \\
    \midrule
    \texttt{lr\_logz} & 1e-1 \\
    \texttt{lr} & 5e-3 \\
    \texttt{lr\_patience} & 10 \\
    \texttt{n\_samples} & 100 \\
    \texttt{top\_n} & 50 \\
    \texttt{batch\_size} & 64 \\
    \texttt{epsilon} & 0.25 \\
    \bottomrule
  \end{tabular}
\end{table}

\begin{table}[htbp]
  \centering
  \caption{Teacher / curriculum selection hyperparameters}
  \label{tab:teacher-hparams}
  \small
  \begin{tabular}{@{}ll@{}}
    \toprule
    Hyperparameter & Value \\
    \midrule
    \texttt{lpe} (learning-progress estimator) & Online \\
    \texttt{acp} (progress metric) & LP \\
    \texttt{a2d} (action-to-difficulty) & GreedyProp \\
    \texttt{a2d\_eps} & 0.15 \\
    \texttt{lpe\_alpha} & 0.05 \\
    \texttt{acp\_MR\_K} & 25 \\
    \texttt{acp\_MR\_power} & 2 \\
    \texttt{acp\_MR\_pot\_prop} & 0.4 \\
    \texttt{acp\_MR\_att\_pred} & 0.1 \\
    \texttt{acp\_MR\_att\_succ} & 0.05 \\
    \bottomrule
  \end{tabular}
\end{table}

\paragraph{Values explored during tuning. } During preliminary tuning, we explored multiple ranges for each major hyperparameter (final values are reported in the tables). For model capacity we tested  \(\texttt{embedding\_dim}\in\{16,32,64\}\), \(\texttt{hidden\_dim}\in\{128,256,512\}\), and \(\texttt{n\_hidden}\in\{2,4,6\}\). For optimization we swept learning rates with \(\texttt{lr}\in\{5\times10^{-4},\,1\times10^{-3},\,5\times10^{-3}\}\) and \(\texttt{lr\_logz}\in\{1\times10^{-2},\,1\times10^{-1}\}\). For the CL algorithm, we varied teacher parameters such as \(\texttt{lpe\_alpha}\in\{0.01,0.05,0.1\}\) and \(\texttt{acp\_MR\_K}\in\{10,25,50\}\).

For each hyperparameter class we ran short preliminary experiments (a small number of curriculum iterations) to evaluate stability, wall-clock cost, and average learning progress. The reported defaults were selected as a trade-off between stable learning dynamics, empirical performance in these exploratory runs, and computational efficiency, when multiple configurations performed similarly we prioritized smaller/cheaper configurations. All experiments in the main text use the values listed in Tables~\ref{tab:curriculum-hparams}--\ref{tab:teacher-hparams}.

\subsection{Curriculum Learning Configurations}
\label{app:cl-configs}
To systematically evaluate the impact of different curriculum learning (CL) strategies on the multi-objective mRNA design task, we designed three distinct experimental configurations. These configurations are tailored to probe the trade-offs between learning stability, adaptiveness, and the exploration-exploitation balance. Each configuration uses a unique combination of components for Learning Progress Estimation (LPE), Attention Computation (ACP), and Attention-to-Distribution mapping (A2D), allowing us to isolate and analyze their effects on model performance.

\paragraph{Configuration 1: Conservative EMA-based Curriculum. }
This configuration is designed to prioritize stable, gradual learning, making it potentially more robust to the high variance often present in biological reward landscapes.
\begin{itemize}
    \item Learning Progress Estimation (LPE): It employs an Online Exponential Moving Average (EMA) (lpe='Online') with a very small smoothing factor ($\alpha$=0.05). This ensures that the progress estimate is robust to noisy rewards and changes slowly, preventing drastic shifts in the curriculum.
    \item Attention Computation (ACP): Task selection relies on a simpler Learning Progress (LP) attention mechanism (acp='LP'), which directly uses the smoothed learning rate to prioritize tasks.
    \item Distribution Mapping (A2D):  It uses an $\epsilon$-greedy proportional strategy (a2d='GreedyProp') with a moderate exploration rate of $\epsilon$=0.15. This forces the model to dedicate 15$\%$ of its time to uniform exploration, ensuring it does not prematurely converge on a suboptimal subset of tasks.
\end{itemize}
\paragraph{Configuration 2: Aggressive Sampling-based Curriculum. }
In contrast, this setup is engineered for rapid adaptation and aggressive exploration. It is intended to quickly discover and exploit promising areas of the vast mRNA design space.
\begin{itemize}
    \item Learning Progress Estimation (LPE): It utilizes a Sampling-based progress estimator (lpe='Sampling') over a small, recent window of performance (K=10). This makes the curriculum highly responsive to immediate learning signals.
    \item Attention Computation (ACP): It leverages a more sophisticated Mastering Rate (MR) attention mechanism (acp='MR') with a high power parameter (power=8) and a strong emphasis on task potential (pot\_prop=0.8). This aggressively prioritizes tasks that show the most promise for rapid improvement.
\end{itemize}
\vspace{-0.5cm}
\paragraph{Configuration 3: Balanced Proportional Curriculum. }
This configuration seeks a robust balance between stability and adaptability, making it hypothetically well-suited for the complex trade-offs inherent in multi-objective optimization.

\begin{itemize}
    \item Learning Progress Estimation (LPE): Learning progress is estimated using Linear Regression (lpe='Linreg') over a larger window (K=25). This provides a stable yet responsive measure of the learning trend, which is effective for handling noisy, multi-objective rewards by capturing the underlying slope of improvement.
    \item Attention Computation (ACP): It also uses the Mastering Rate (MR) attention mechanism but with moderate parameters (power=4, pot\_prop=0.6) to balance focus between currently improving tasks and potentially valuable new ones.
    \item Distribution Mapping (A2D): Task distribution is purely proportional (a2d='Prop') to the computed attention weights. This represents a pure exploitation strategy based on the curriculum's policy, with no additional random exploration ($\epsilon$=0.0).
\end{itemize}

% \section{Curriculum Learning Algorithm}
% \label{app:cl_algorithm}

\begin{figure}[H]
    \centering
    \begin{subfigure}[b]{0.48\linewidth}
        \centering
        \includegraphics[width=\linewidth]{ 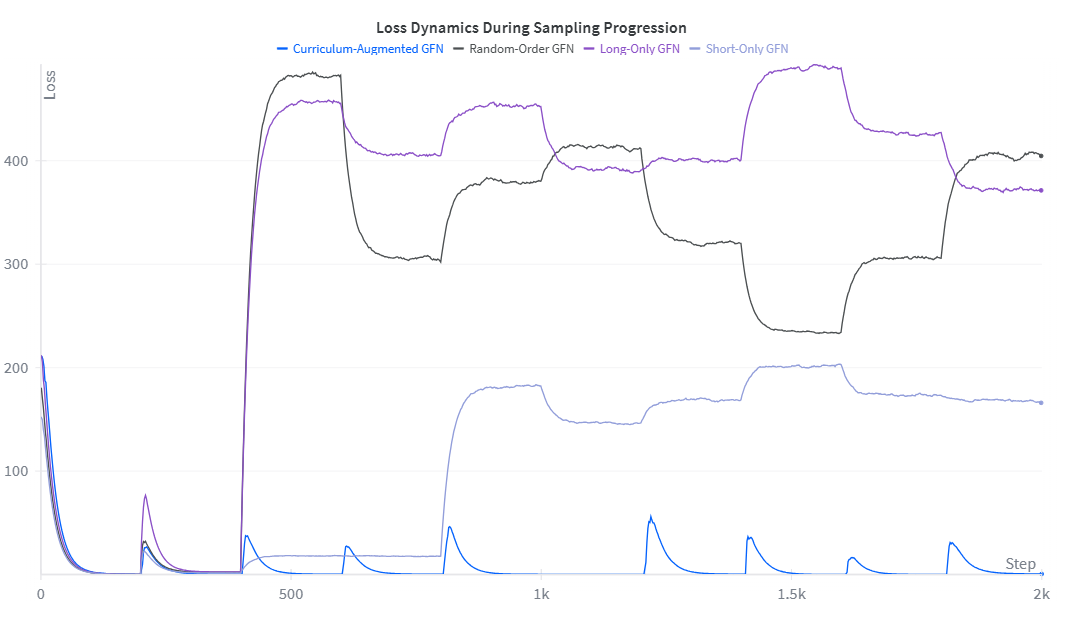}
        \caption{Training loss across the four regimes.}
        \label{fig:loss_dynamics_a}
    \end{subfigure}
    \hfill
    \begin{subfigure}[b]{0.48\linewidth}
        \centering
        \includegraphics[width=\linewidth]{ 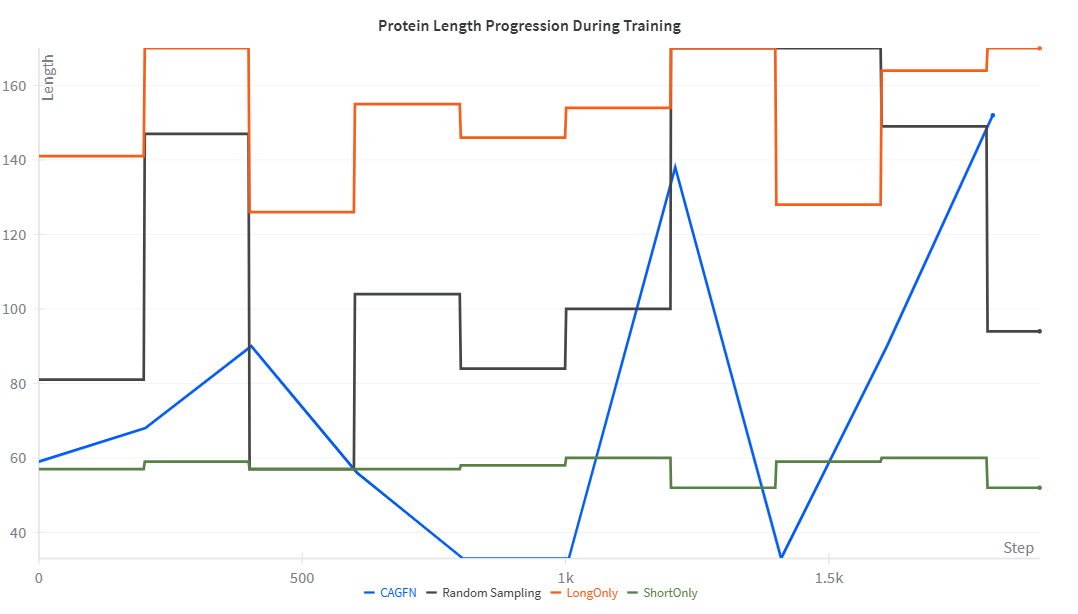}
        \caption{Protein sequence length distribution.}
        \label{fig:loss_dynamics_b}
    \end{subfigure}
    \caption{Loss dynamics and curriculum progression. (a) Training loss for \textit{Short-Only}, \textit{Long-Only}, \textit{Random-Order}, and \textit{Curriculum-Augmented GFN (ours)}. (b) Corresponding progression of protein sequence lengths shown to the model during training.}
    \label{fig:loss_dynamics}
\end{figure}

In figure~\ref{fig:loss_dynamics} : Panel (a) shows that the three non-curriculum baselines (\textit{Short-Only}, \textit{Long-Only}, \textit{Random-Order}) fail to consistently reduce the loss and do not converge across training steps.

\section{Additional Plots}
\label{app:additional-plots}

\begin{figure}[H]
    \centering
    \begin{subfigure}{0.73\textwidth}
        \centering
        \includegraphics[width=\linewidth]{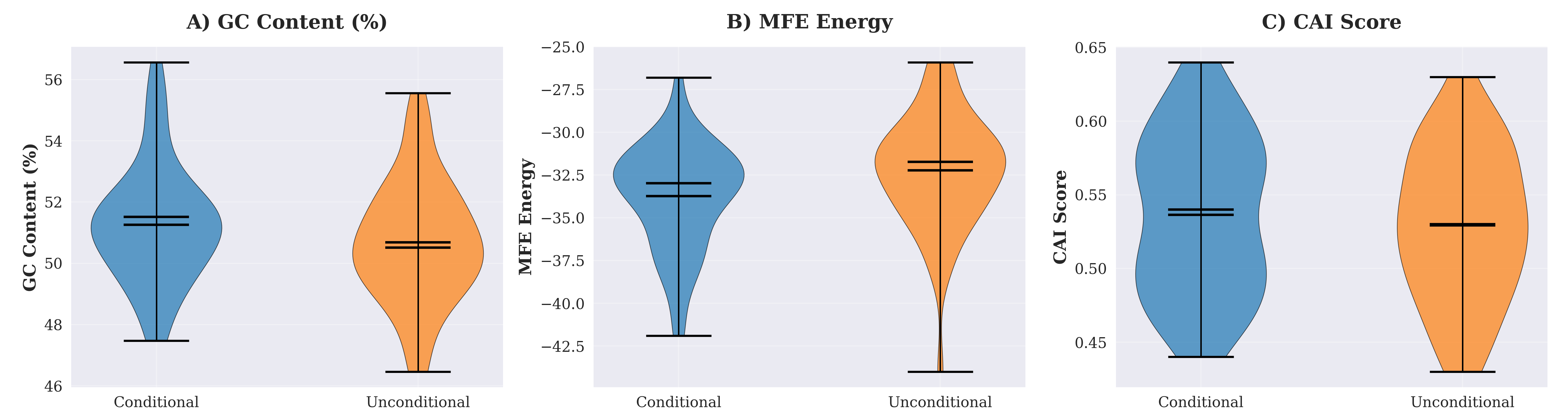}
        \label{fig:condi_small}
    \end{subfigure}
    \hfill
    \begin{subfigure}{0.23\textwidth}
        \centering
        \includegraphics[width=\linewidth]{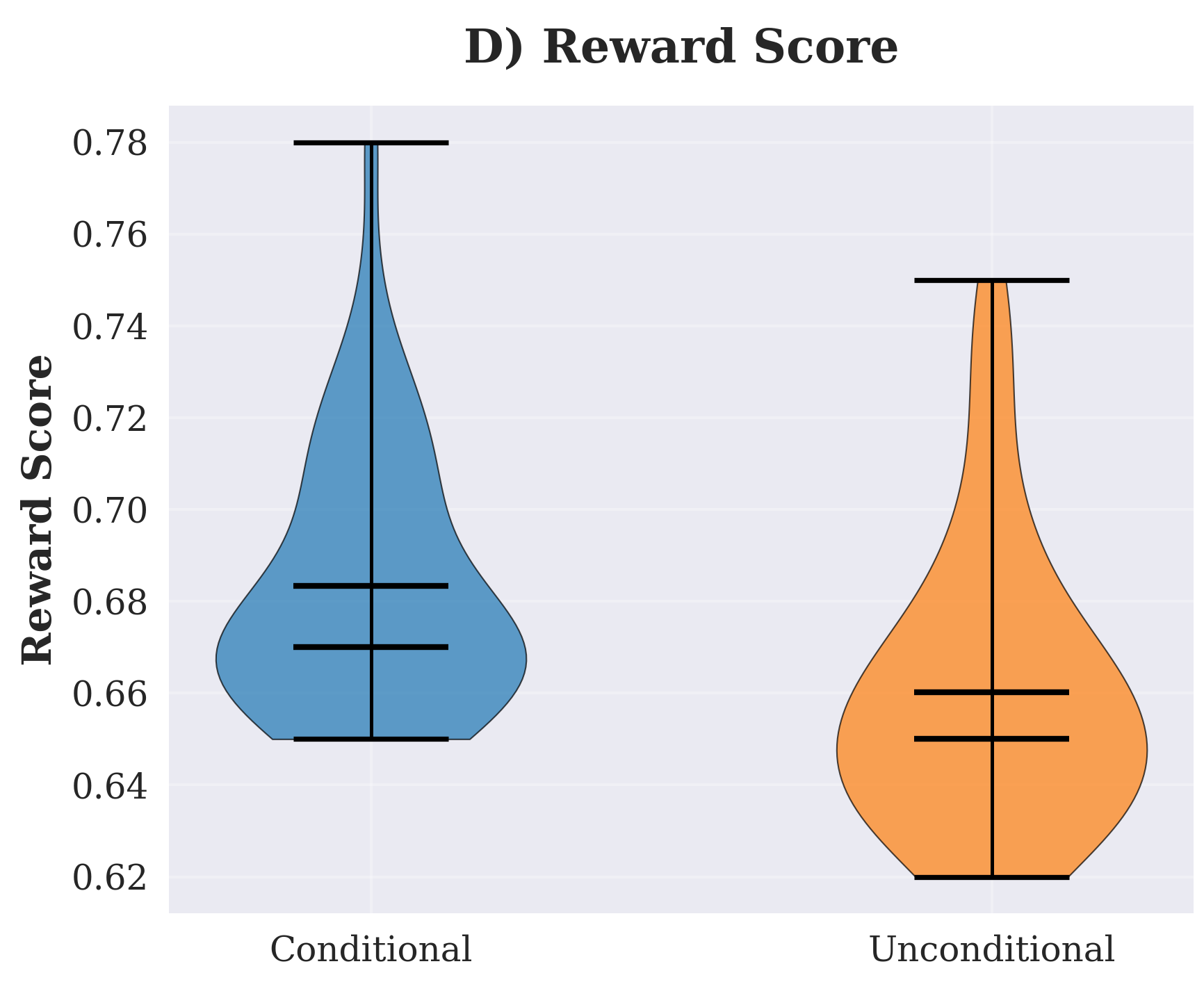}
        \label{fig:condi_reward}
    \end{subfigure}
    \caption{\textbf{Conditional vs. unconditional mRNA generation results.}
    Metrics distribution across mRNA sequences for a small target protein ($\sim$35 AA). \textbf{GC percentage:} higher is better, \textbf{MFE:} lower is better, \textbf{CAI:} higher is better.}
    \label{fig:condi_vs_uncondi}
\end{figure}

\begin{figure}[H]
  \centering

  % top row
  \begin{subfigure}[b]{0.48\linewidth}
    \centering
    \includegraphics[width=\linewidth]{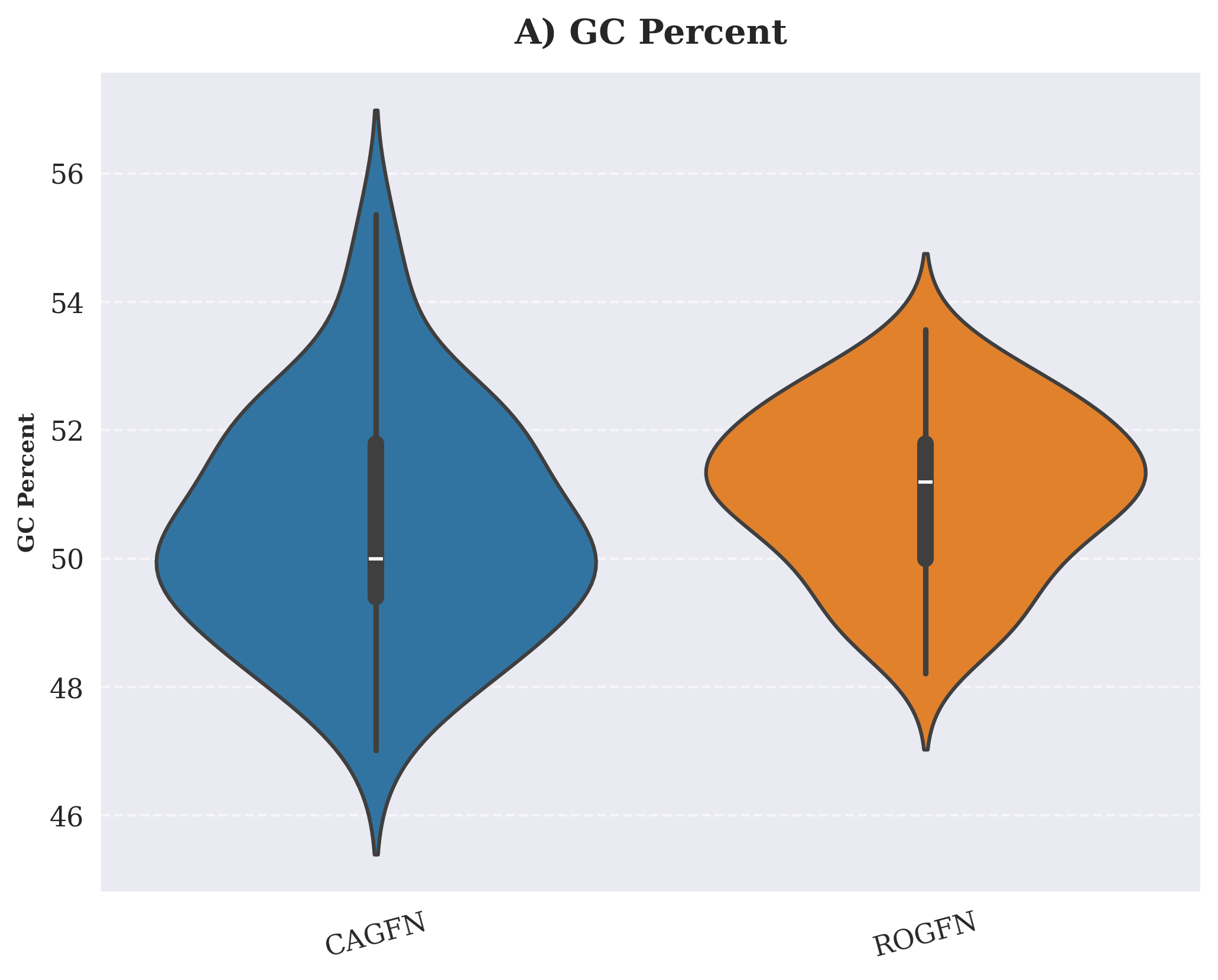}
    \label{fig:medium_gc}
  \end{subfigure}
  \hfill
  \begin{subfigure}[b]{0.48\linewidth}
    \centering
    \includegraphics[width=\linewidth]{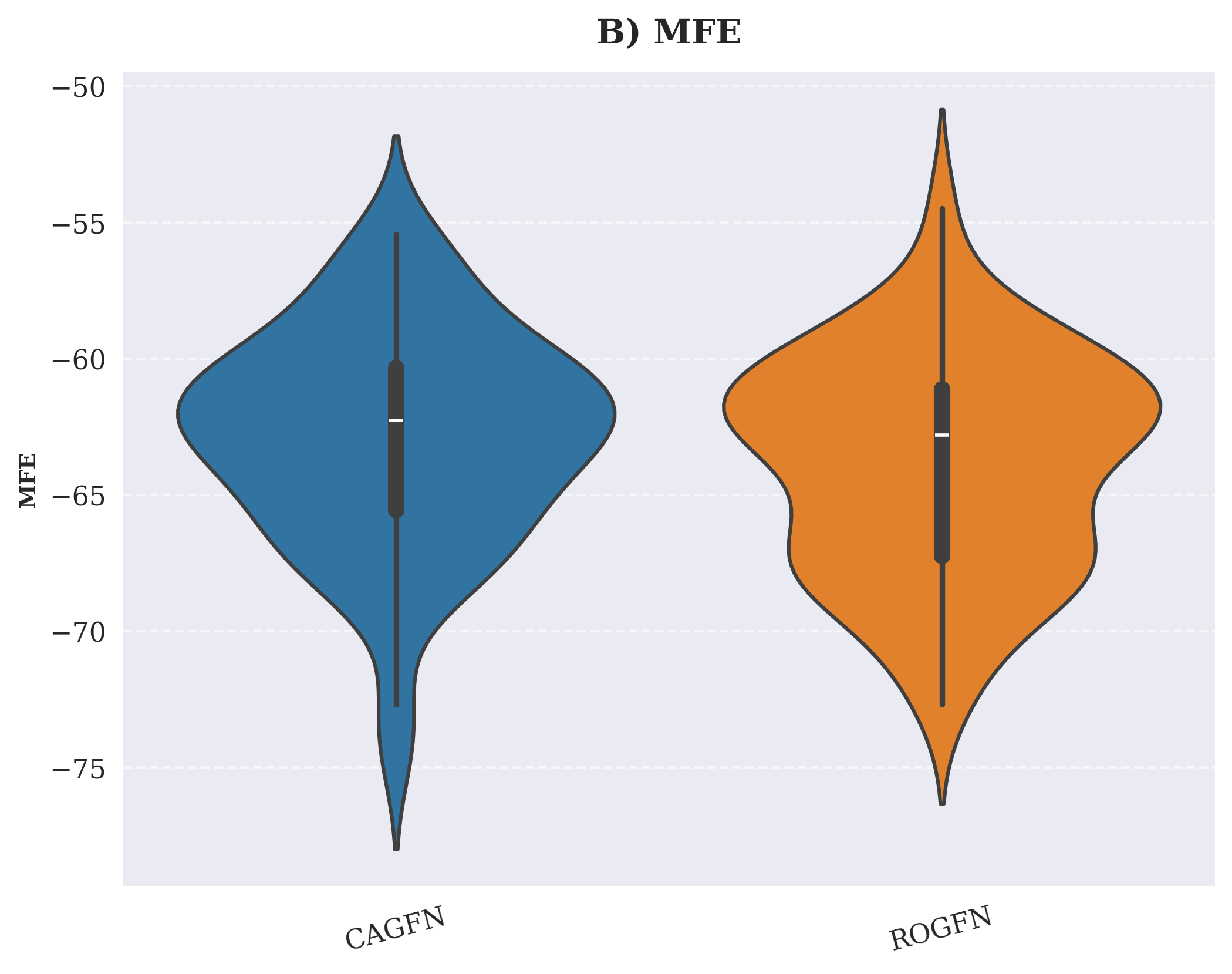}
    \label{fig:medium_mfe}
  \end{subfigure}

  \vspace{4pt}

  % bottom row
  \begin{subfigure}[b]{0.48\linewidth}
    \centering
    \includegraphics[width=\linewidth]{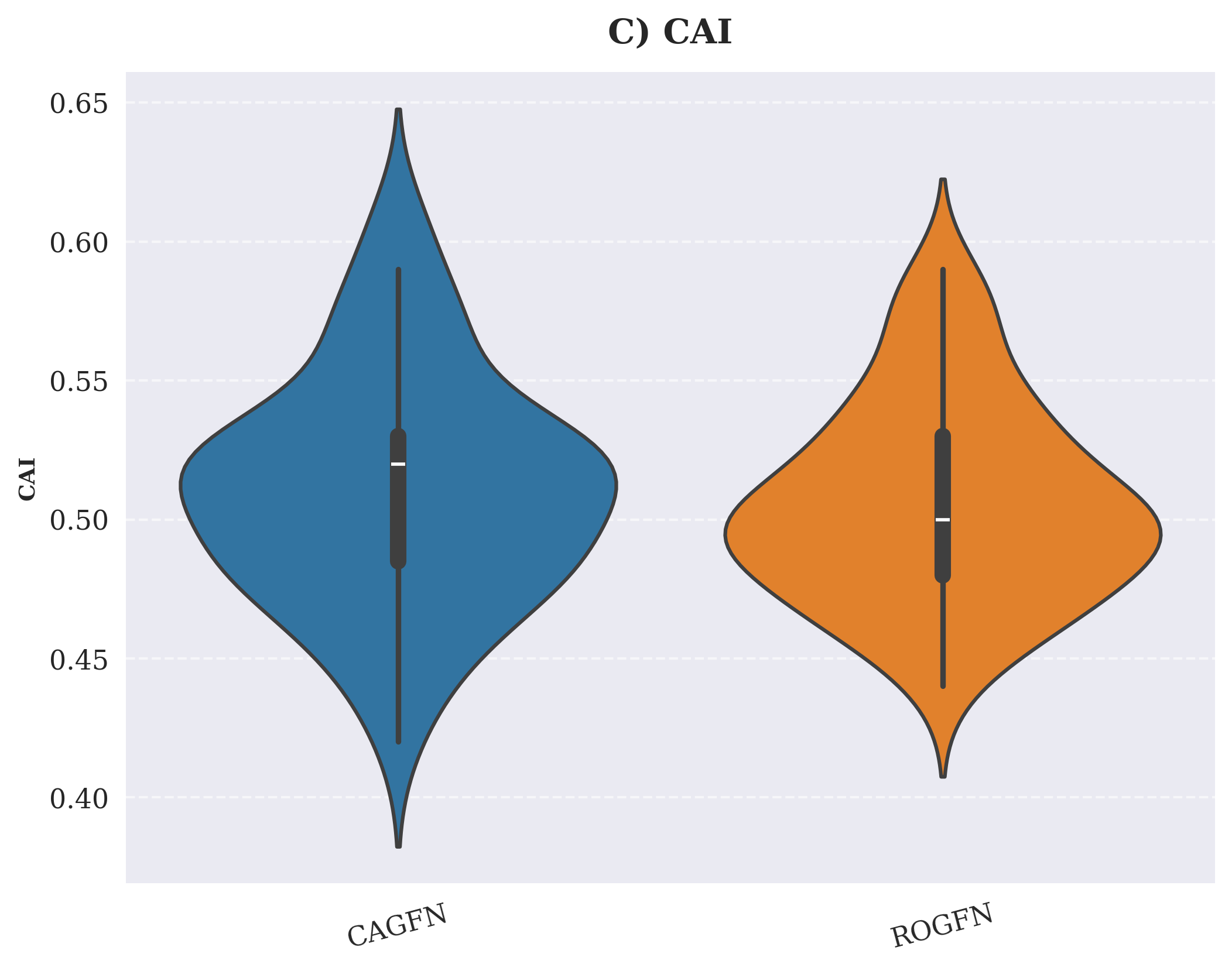}
    \label{fig:medium_cai}
  \end{subfigure}
  \hfill
  \begin{subfigure}[b]{0.48\linewidth}
    \centering
    \includegraphics[width=\linewidth]{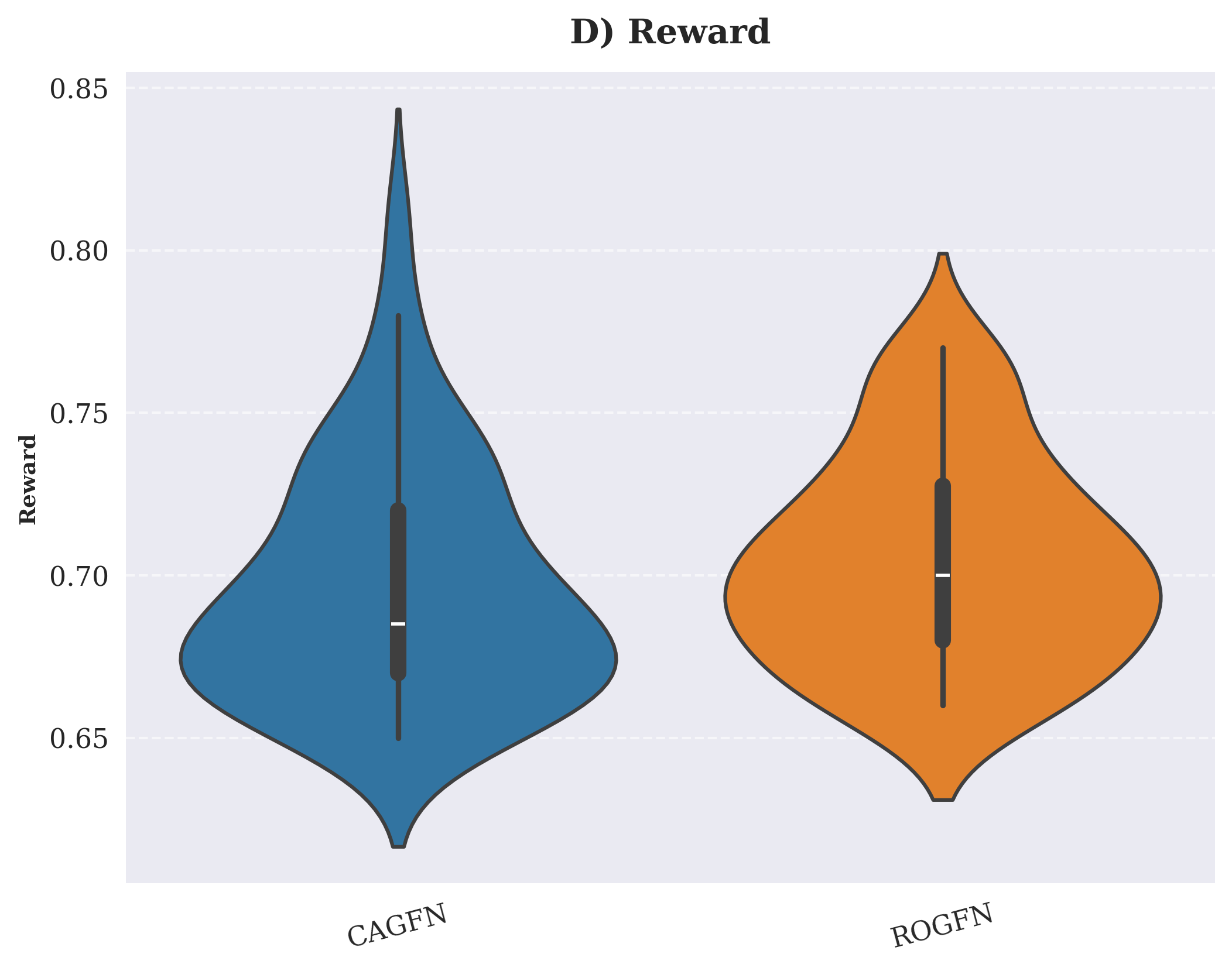}
    \label{fig:medium_reward}
  \end{subfigure}

  \caption{Distribution of reward and objectives for the medium-protein task, contrasting the curriculum-augmented GFlowNet (CAGFN) with a baseline GFlowNet trained without curriculum (ROGFN). Panels show empirical distributions of (a) Reward, (b) CAI, (c) MFE (lower is better), and (d) GC content. CAGFN shifts distributions toward higher reward and CAI, lower MFE, and GC content closer to the desired range.}
  \label{fig:medium}
\end{figure}

\begin{figure}[H]
    \centering
    \begin{subfigure}[b]{0.3\textwidth}
        \centering
        \includegraphics[width=\textwidth]{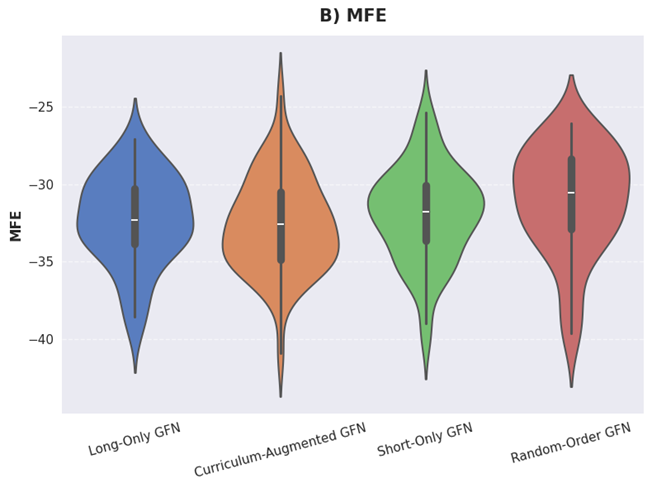}
        \caption{MFE (Low is better)}
    \end{subfigure}
    \hspace{0.02\textwidth}
    \begin{subfigure}[b]{0.3\textwidth}
        \centering
        \includegraphics[width=\textwidth]{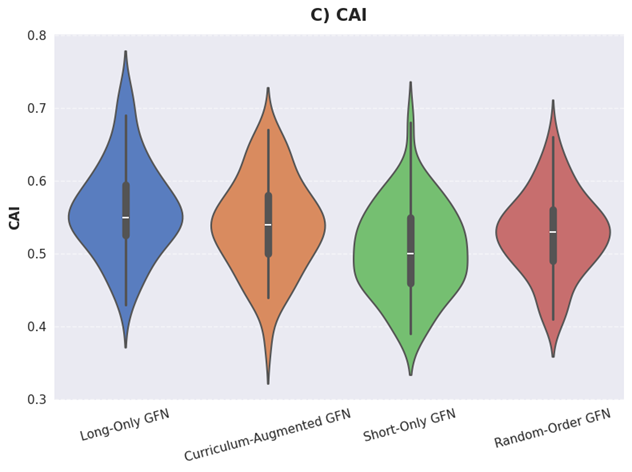}
        \caption{CAI (High is better)}
    \end{subfigure}
    \hspace{0.02\textwidth}
    \begin{subfigure}[b]{0.3\textwidth}
        \centering
        \includegraphics[width=\textwidth]{ 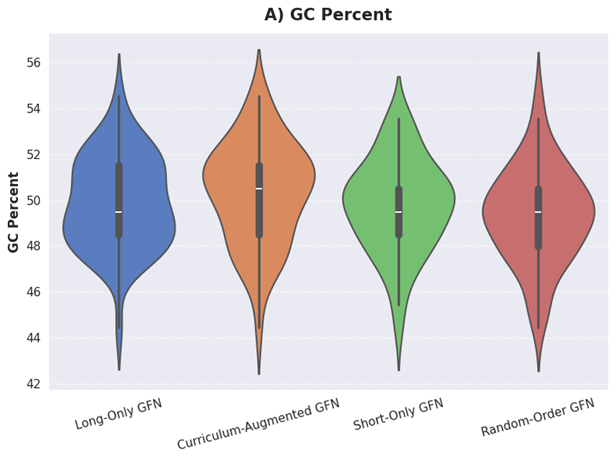} 
        \caption{GC $\%$ (High is better)}
    \end{subfigure}
    \caption{Metrics distribution for the OAZ2 protein}
    \label{fig:seq_metrics}
\end{figure}

\subsection{Loss Curves}
\label{app:loss-curves}
We found that the Sub-Trajectory Balance (SubTB) loss function is critical for maintaining numerical stability for sequences longer than approximately 55 amino acids, as the standard Trajectory Balance (TB) loss consistently failed. All subsequent models are built upon this finding.

\begin{figure}[H]
    \centering
    \begin{subfigure}[b]{0.32\linewidth}
        \centering
        \includegraphics[width=\linewidth]{ 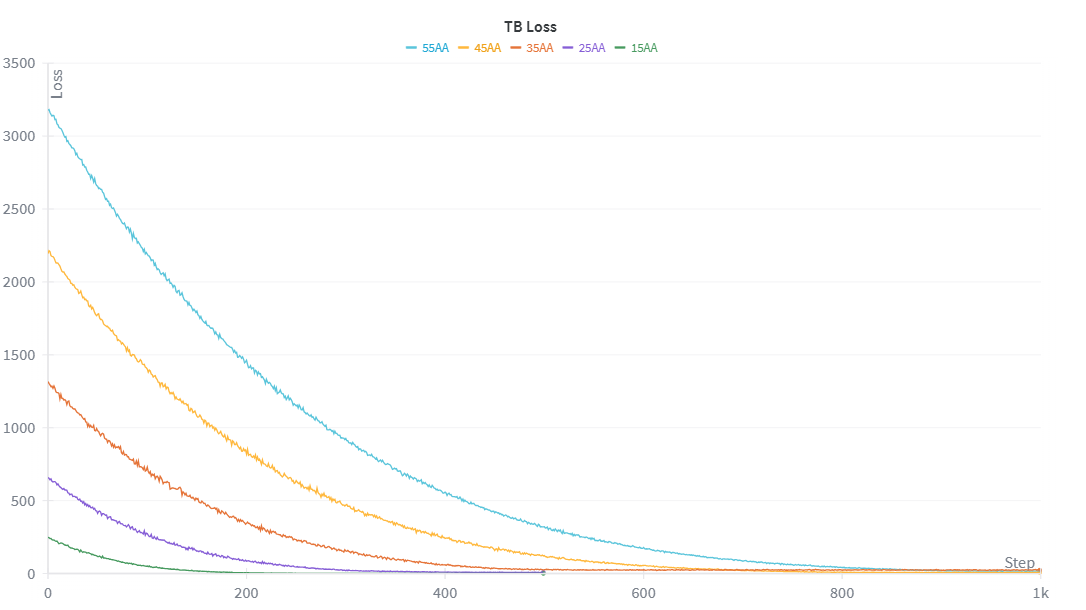}
       \caption{Trajectory Balance (TB) loss on small sequences ($<$75AA).}
        \label{fig:tb_loss1}
    \end{subfigure}
    \hfill
    \begin{subfigure}[b]{0.32\linewidth}
        \centering
        \includegraphics[width=\linewidth]{ 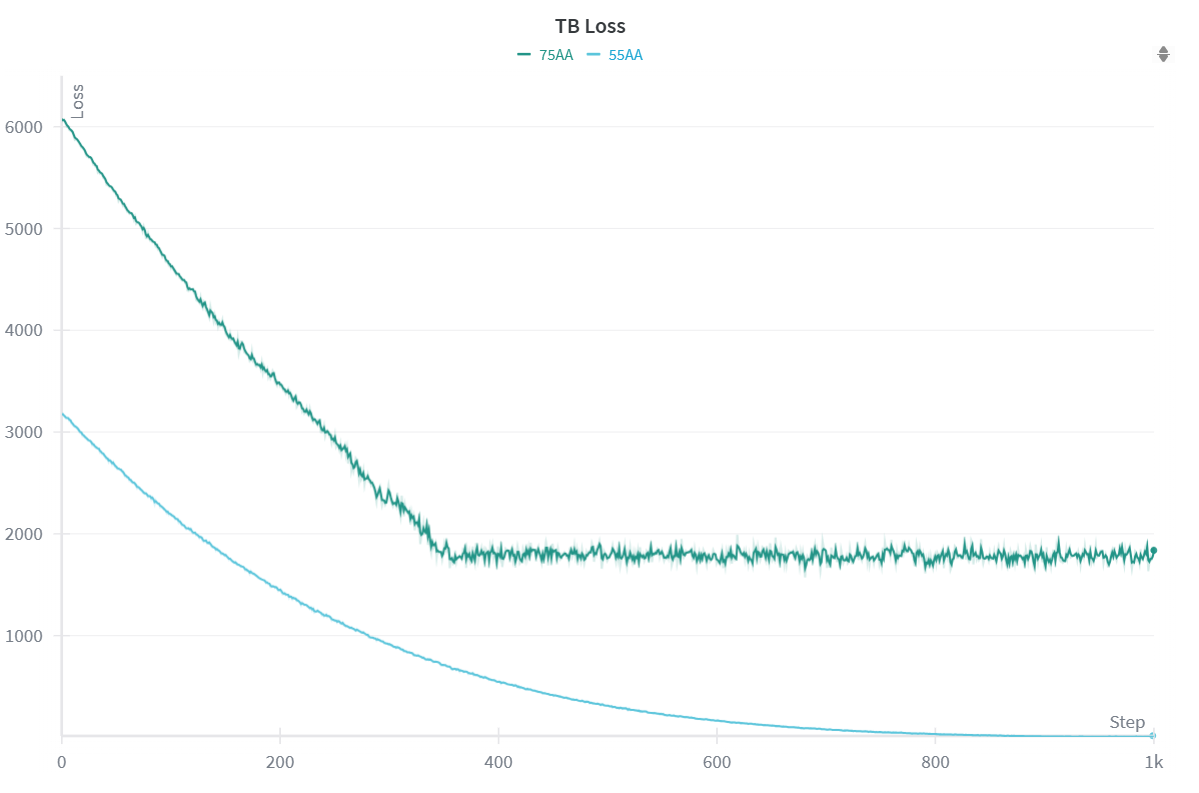}
        \caption{Trajectory Balance (TB) loss on a 75AA sequence.}
        \label{fig:tb_loss2}
    \end{subfigure}
    \hfill
    \begin{subfigure}[b]{0.32\linewidth}
        \centering
        \includegraphics[width=\linewidth]{ 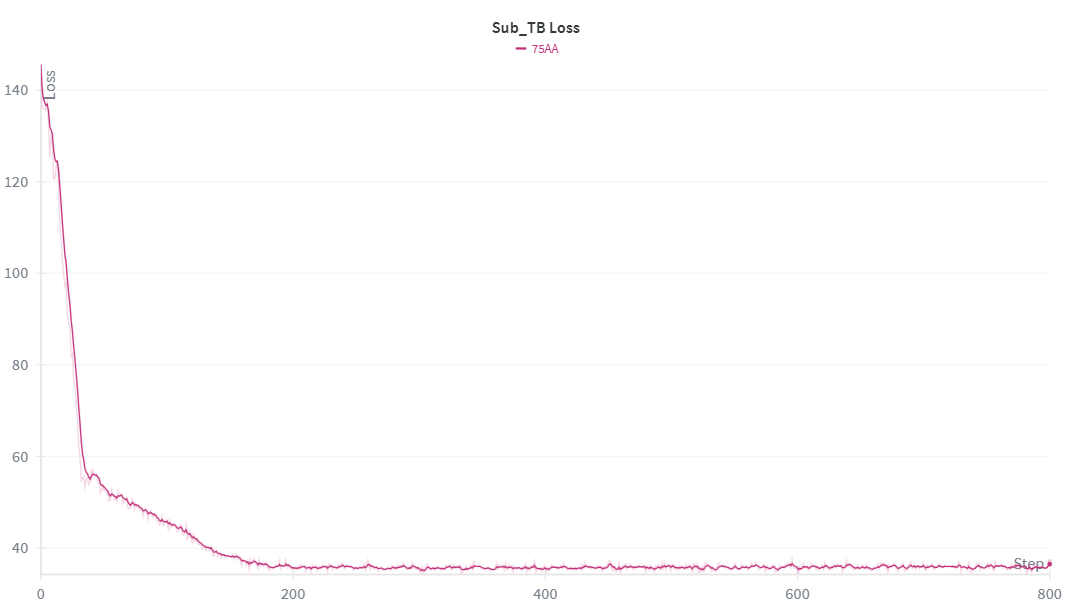}
        \caption{SubTB loss (75AA). Converges stably.}
        \label{fig:subtb_loss}
    \end{subfigure}
    \caption{Sub-Trajectory Balance (SubTB) loss on 75AA sequences. Unlike TB, SubTB converges smoothly and remains numerically stable.}
    \label{fig:loss_comparison}
\end{figure}

We provide additional loss curves in Figures~\ref{fig:tb_loss1}--\ref{fig:subtb_loss} that compare the standard Trajectory Balance (TB) loss with the Sub-Trajectory Balance (SubTB) loss on sequences of length up to 75 amino acids (AA). The results clearly illustrate that TB fails to converge, with the loss plateauing at high values and exhibiting unstable behavior. In contrast, SubTB consistently converges to a stable solution, demonstrating its robustness for longer sequences. 

Furthermore, we investigated whether the convergence issues with TB could be alleviated by changing model architectures or tuning hyperparameters. Specifically, we trained the model using MLP, LSTM, and Transformer architectures across a wide range of learning rates, batch sizes, and hidden dimensions. Across all settings, the TB loss exhibited the same instability, confirming that the problem is inherent to the TB formulation rather than to the architectural choice. These findings motivated our decision to adopt SubTB for all subsequent experiments.

\begin{figure}[H]
    \hfill
    \begin{subfigure}{0.3\linewidth}
        \centering
        \includegraphics[width=\linewidth]{ Protein_Progression.png}
        \caption{}
        \label{fig:progression}
    \end{subfigure}
    \hfill
    \begin{subfigure}{0.3\linewidth}
        \centering
        \includegraphics[width=\linewidth]{ 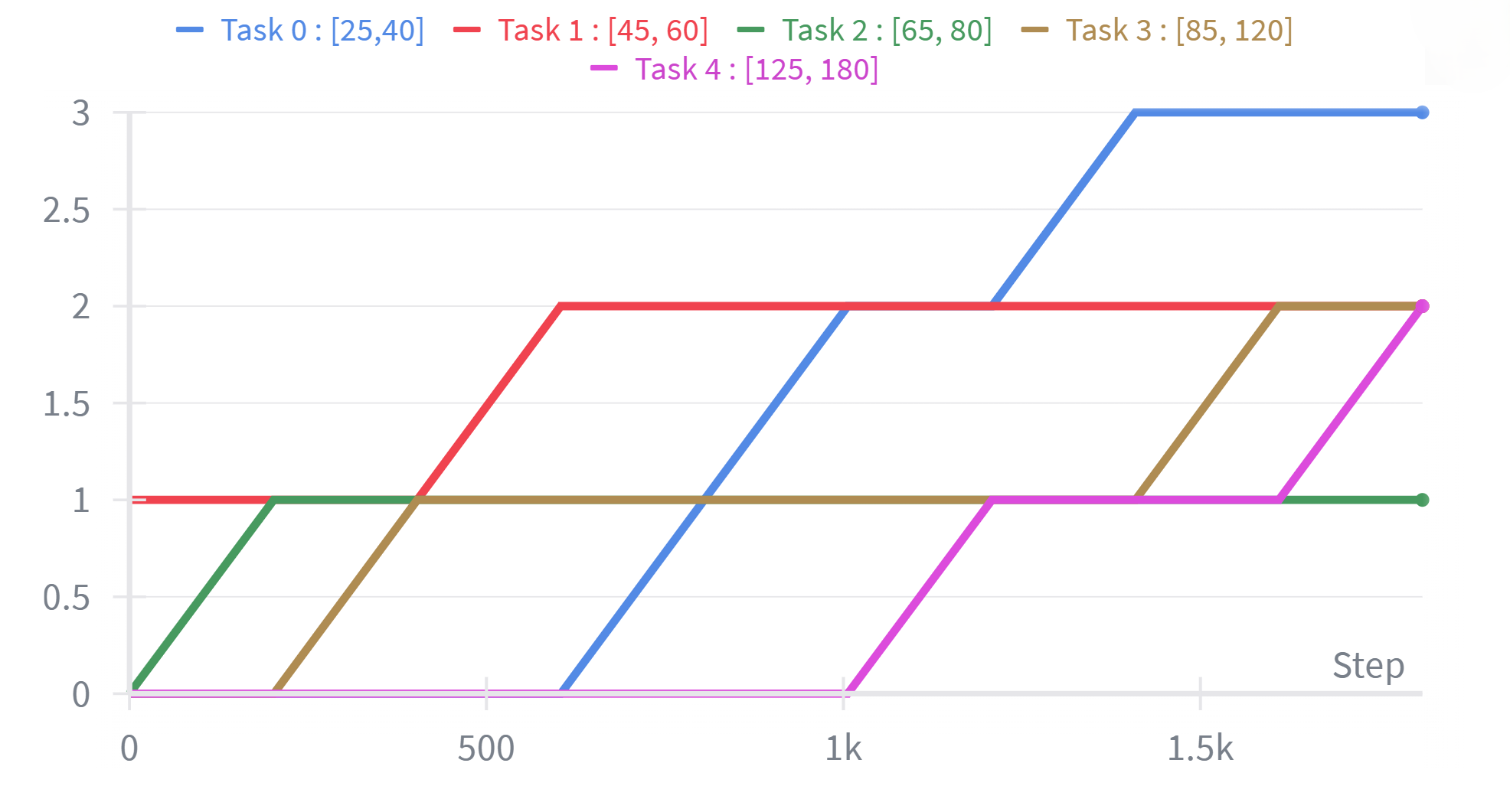}
        \caption{}   
        \label{fig:tasks}
    \end{subfigure}
    \hfill
    \begin{subfigure}{0.3\linewidth}
        \centering
        \includegraphics[width=\linewidth]{ 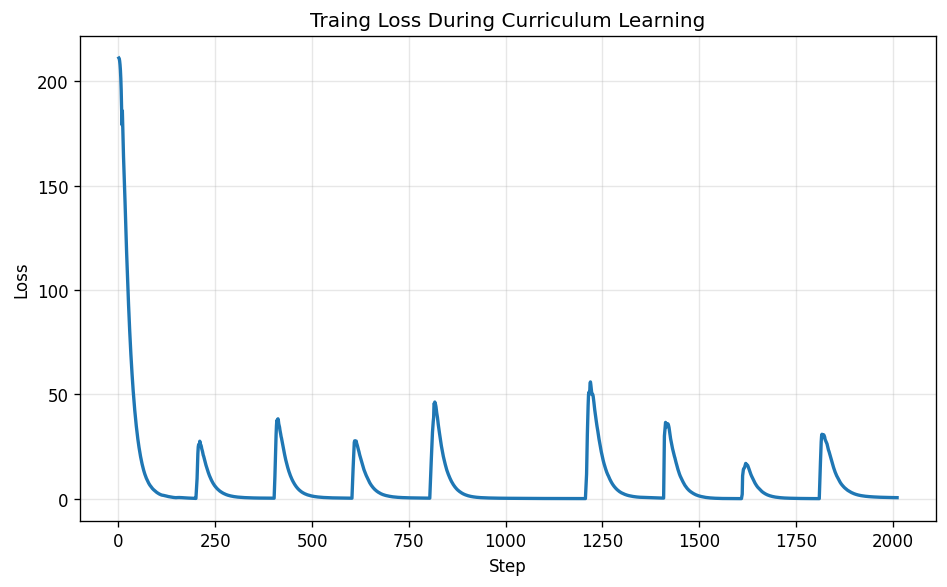}
        \caption{}   
        \label{fig:clloss}
    \end{subfigure}
    
    \caption{\textbf{Curriculum learning dynamics. }(a) We report the protein length progression of the four GFN variants. The comparison shows how the curriculum adapts the sequence length during training compared to alternative strategies. This curriculum schedule reduces the difficulty gap between tasks and allows the model to build competence over time without compromising simple tasks. 
    (b) Indicates how often tasks of different lengths are sampled during curriculum, it shows the balance between exploration across easy and hard tasks, ensuring that the training process remains stable while covering the full range of sequence lengths. (c) CAGFN Loss. Brief spikes or oscillations may occur around curriculum tasks transitions.}
    \label{fig:curriculum_overview}
\end{figure}

\section{Extended related work}
\label{app:related-work}

\paragraph{GFlowNets for scientific design.}
GFlowNets learn stochastic policies that sample compositional objects with probability proportional to a reward function, intrinsically promoting diversity over mode collapse \citep{bengio2021flow, Bengio2023GFlowNet}. 
\citet{Jain2022GFlowNet} embed a GFlowNet in an active learning loop to generate diverse candidate peptides, finding more novel high-scoring sequences than previous methods. In drug discovery, Reaction-GFN~\citep{koziarski2024rgfn} constrains generation to chemical reaction steps so that every molecule has a valid synthetic route, and SynFlowNet~\citep{Cretu2024SynFlowNet} restricts actions to documented reactions and purchasable starting materials to produce synthetically accessible molecules while preserving diversity. These applications demonstrate that GFlowNets can flexibly incorporate domain-specific action spaces to generate complex, structured objects.

\paragraph{Multi-objective GFlowNets.}
Many design problems involve multiple, often conflicting objectives.
\citet{MOGFN2023} propose Multi-Objective GFlowNets (MOGFNs) to directly sample diverse Pareto-optimal solutions, achieving improved Pareto-front coverage and candidate diversity compared to standard RL baselines. 
\citet{zhu2023sample} integrate GFlowNets into a multi-objective Bayesian optimization framework via a hypernetwork-based GFlowNet (HN-GFN) that generates batches of molecules across different trade-offs, effectively sampling an approximate Pareto front. These works show that GFlowNets can be conditioned on objective preference vectors to explore trade-off spaces rather than collapsing to a single optimum.

\paragraph{Existing approaches to mRNA design.}
The dominant computational approach to mRNA sequence design is LinearDesign \citep{zhang2023algorithm}, which reduces the codon optimization problem to lattice parsing and finds sequences that jointly optimize structural stability (MFE) and codon usage (CAI) in near-linear time. While highly efficient, LinearDesign returns a single optimal or near-optimal sequence per run and does not natively explore trade-offs across multiple objectives or generate diverse candidate pools. More recently, CodonTransformer \citep{Fallahpour2025CodonTransformer} uses a transformer-based language model trained on natural coding sequences to predict codon choices, but it similarly produces point predictions rather than diverse sets of candidates. 
To our knowledge, no prior work has applied GFlowNets to mRNA design. The task introduces distinct challenges compared to small-molecule generation: codon redundancy creates an exponentially large design space, biological constraints (stability, translational efficiency, GC content) must be balanced simultaneously, and the resulting sequences are substantially longer than typical molecular graphs.

\paragraph{Curriculum learning.}
Curriculum learning (CL) structures training so the learner progresses from easier to harder tasks \citep{bengio2009curriculum}. Self-paced learning \citep{kumar2010self} extends this idea by letting the model itself determine difficulty based on its current loss, dynamically selecting easy samples at each iteration.
In the RL setting, Teacher--Student Curriculum Learning \citep[TSCL;][]{matiisen2019teacher} formalizes this with a teacher that monitors per-task learning curves and allocates training to tasks where the student shows the fastest progress. \citet{willems2020mastering} refine TSCL by replacing learning progress with a \emph{mastering rate} signal, which accounts for both current competence and remaining potential, yielding improved sample efficiency.
Despite its success in RL, curriculum learning has not been combined with GFlowNets. The long horizons and sparse terminal rewards of mRNA sequence generation make it a natural candidate for curriculum-based training, which we validate in this work.

%Taken together, these strands motivate our approach. We adopt the MOGFN formulation as the base model for our curriculum-augmented GFlowNet (CAGFN) to (i) generate diverse, Pareto-aware mRNA candidates and (ii) exploit a teacher-driven curriculum that progressively adapts sequence length during training based on measured learning progress. This combination addresses both the multi-objective nature of mRNA design and the training challenges posed by the length of these sequences.

\paragraph{Summary.}
We adopt the MOGFN formulation as the base model for our curriculum-augmented GFlowNet (CAGFN) to (i) generate diverse, Pareto-aware mRNA candidates and (ii) exploit a teacher-driven curriculum that progressively adapts sequence length during training based on measured learning progress. This combination addresses both the multi-objective nature of mRNA design and the training challenges posed by the length of these sequences.

\section{mRNA Codon Design Environment}
\label{sec:codondesignenv}
To exploit codon redundancy in protein coding, we introduce the \textsc{\textbf{CodonDesignEnv}}, a discrete environment for codon-level mRNA design given a protein sequence. The environment models mRNA construction as a sequential decision process: states represent partially constructed sequences, actions correspond to choosing codons, and trajectories terminate once a full sequence encoding the target protein is completed.

\paragraph{State Space ($\mathcal{S}$). }  
Each state $s \in \mathcal{S}$ is represented as a fixed-length integer vector of codon indices of size $L$ (protein length). Unassigned positions are denoted by $-1$, and the initial state is
\[
s_0 = (-1, -1, \dots, -1).
\]
A state becomes terminal once all $L$ positions are filled with valid codons, yielding a complete mRNA sequence $x$.

\paragraph{Action Space. }  
The action set consists of all $64$ codons plus a special ``exit'' action $a_{\text{exit}}$, i.e. $|\mathcal{A}| = 65$. At step $t$, the model selects a codon for the $t$-th amino acid in the target protein. Upon completion ($t = L$), the only valid action is $a_{\text{exit}}$, which transitions the environment to a terminal sink state $s_f$.

\paragraph{Dynamic Masking Strategy. }  
To ensure biological validity, the environment uses dynamic action masking:
\begin{itemize}
    \item \textbf{Synonymous codon masking:} At step $t<L$, the forward action set $\mathcal{A}_f(s)$ is restricted to codons encoding the $(t+1)$-th amino acid.  
    \item \textbf{Termination masking:} At step $t=L$, only $a_{\text{exit}}$ is permitted, forcing proper termination.  
    \item \textbf{Backward masking:} For backward sampling, only the most recently added codon can be removed, preserving sequential construction.  
\end{itemize}
These constraints define a directed acyclic graph (DAG) where each path from $s_0$ to $s_f$ corresponds to a valid mRNA sequence.

\paragraph{Reward Function. } 
\label{reward:conditioning}
Rewards are assigned only to terminal states and capture biologically motivated objectives (\S\ref{mrna_objectives}): GC content, minimum free energy (MFE), and codon adaptation index (CAI). For a complete mRNA sequence $x$, the reward is
\[
R(x) = w^\top \phi(x),
\]
where $\phi(x)$ is a vector of normalized objectives and $w \in \mathbb{R}^3$ are user-defined weights. By adjusting $w$, one can bias the GFlowNet toward different Pareto-optimal trade-offs, promoting diversity and mode coverage across the design space.

While we adopt the weighted-sum formulation in this work for simplicity and interpretability, the same framework is compatible with alternative scalarization schemes. More generally, our approach follows the idea of \emph{preference-conditional GFlowNets}  \citep[MOGFN-PC;][]{MOGFN2023}, in which a single reward-conditional GFlowNet models the entire family of multi-objective optimization sub-problems simultaneously. Here, preferences $\omega \in \Delta^d$ over the set of objectives $\{R_1(x), \dots, R_d(x)\}$ act as conditioning variables, and the reward function is defined via a scalarization $R(x \mid \omega)$. This general formulation accommodates any scalarization function, whether classical weighted sums, max–min formulations, or novel functions designed for specific biological contexts. For example, alternatives such as the weighted-log-sum scalarization have been proposed to mitigate cases where one objective dominates the reward signal.

\end{document}